\pdfoutput=1
\documentclass[11pt]{article}
\PassOptionsToPackage{hyphens}{url}
\usepackage[final]{acl}
\usepackage{times}
\usepackage{latexsym}
\usepackage[T1]{fontenc}
\usepackage[utf8]{inputenc}
\usepackage{microtype}
\usepackage{inconsolata}
\usepackage{graphicx}
\usepackage{amsmath}
\usepackage{bbm}
\usepackage{hyperref}
\usepackage{booktabs}
\usepackage{multirow}
\usepackage{tikz}
\usepackage{amssymb}
\usetikzlibrary{positioning,arrows.meta,calc,fit,shapes.geometric,backgrounds}
\usepackage{tabularx}
\usepackage[most]{tcolorbox}
\usepackage{enumitem}
\usepackage{stfloats}
\usepackage{placeins}
\usepackage{algorithm}
\usepackage{algpseudocode}

\usepackage[table]{xcolor}
\usepackage{array}

\definecolor{PanelFill}{HTML}{EEF2F6}
\definecolor{MedRed}{HTML}{B14A4A}

\newcolumntype{C}[1]{>{\centering\arraybackslash}p{#1}}

\newcommand{\best}[1]{\textbf{#1}}
\newcommand{\second}[1]{\underline{#1}}

\DeclareRobustCommand{\medmark}{%
  \textcolor{MedRed}{\raisebox{0.05ex}{\scriptsize\textsf{\textbf{+}}}}%
}
\title{Dense Clinical Contrasts Enhance Medical Knowledge Updating in Large Language Models}

\author{
  \textbf{Yangmin Huang}\textsuperscript{1,2,*,\dag},
  \textbf{Shu Quan}\textsuperscript{1,*},
  \textbf{He Geng}\textsuperscript{1},
  \textbf{Xin Ye}\textsuperscript{1,\dag},
  \textbf{Qianyun Du}\textsuperscript{1},
\\
  \textbf{Zhiyang He}\textsuperscript{1},
  \textbf{Jiaxue Hu}\textsuperscript{1},
  \textbf{Xiaodong Tao}\textsuperscript{1}
\\
  \textsuperscript{1} Xunfei Healthcare Technology Co., Ltd.
\\
  \textsuperscript{2} USTC-Xunfei Healthcare Digital and Health Joint Laboratory
\\
  \texttt{\{ymhuang9, shuquan, hegeng2, xinye6, qydu, zyhe, jxhu2, xdtao\}@iflytek.com}
}

\begin{document}
\maketitle

\renewcommand{\thefootnote}{\fnsymbol{footnote}}
\footnotetext[1]{Equal contribution.}
\footnotetext[2]{Corresponding authors.}
\renewcommand{\thefootnote}{\arabic{footnote}}

\begin{abstract}
Medical knowledge changes continually, making large language models vulnerable to relying on outdated yet clinically plausible information. We study whether the format of supervision affects medical knowledge updating under a matched training-budget setting. We introduce SEER-Bench, a temporally anchored oncology-staging benchmark curated from the latest versioned SEER Research Data release, and render identical medical update events from NCCN oncology guidelines into four supervision formats: \textsc{EMQ}, \textsc{MSQ}, \textsc{FITB}, and \textsc{SAQ}. Across SEER-Bench and HealthBench Professional, \textsc{EMQ} gives the most stable external transfer and retention among same-budget SFT variants. With \textsc{EMQ} supervision, the updated 4B model produces competitive results on temporally anchored oncology staging, reaching 64.8\% answer accuracy and 59.6\% rationale accuracy on SEER-Bench. Diagnostic analyses suggest that \textsc{EMQ} exposes denser clinical contrast signals while preserving discriminative representations with smaller movement from the base model. These results show that medical knowledge updating depends not only on the update algorithm, but also on how knowledge is structured as supervision.
\end{abstract}
\section{Introduction}

Large language models (LLMs) are increasingly used for medical question answering and clinical decision support~\citep{thirunavukarasu2023large,singhal2023large,nori2023capabilities}, yet their parametric knowledge can become stale as medical standards update~\citep{lazaridou2021mind,dhingra-etal-2022-time}. This challenge is especially critical in oncology, where staging criteria, treatment guidelines, and drug labels are frequently revised~\citep{strobl2023treatment}. These revisions are often incremental and highly conditional; even narrowly scoped guideline revisions can substantially alter clinical decision pathways. For instance, the latest NCCN updates prioritize immediate EBRT or systemic therapy for unresectable, locoregional invasive thyroid cancer, rendering the previous routine standard of $^{131}$I whole-body imaging obsolete~\citep{nccn2026thyroid}. When such updates are not incorporated, LLMs may assign an outdated stage that changes eligibility for surgery, radiotherapy, or systemic treatment; recommend a superseded management pathway that delays appropriate care; or support an obsolete recommendation with an overconfident rationale that makes the error difficult to detect; Appendix~\ref{sec:appendix-clinical-safety} gives a versioned guideline example of each potential failure mode. These failures may occur unevenly across cancer populations, particularly when certain cancer types, clinical presentations, languages, or guideline regions are underrepresented in training or update supervision.

Although various methods address LLM knowledge staleness—ranging from retrieval to parametric editing~\citep{wang2024easyedit}—one question remains underexplored in temporally evolving medical adaptation: how should knowledge updates be represented as supervision before adaptation? Moreover, systematically evaluating how these supervision formats affect model adaptation requires an explicit temporal anchor to determine whether performance reflects pre-existing parametric knowledge or the successful incorporation of newly revised guidance. Yet existing medical benchmarks primarily assess long-stable textbook knowledge~\citep{jin2021disease,pal2022medmcqa}, while temporal benchmarks focus on general-domain facts~\citep{vu2024freshllms}, leaving a gap for a benchmark with a well-defined clinical time boundary for evaluating period-specific medical updates.

\begin{figure*}[t]
    \centering
    \includegraphics[width=1\linewidth]{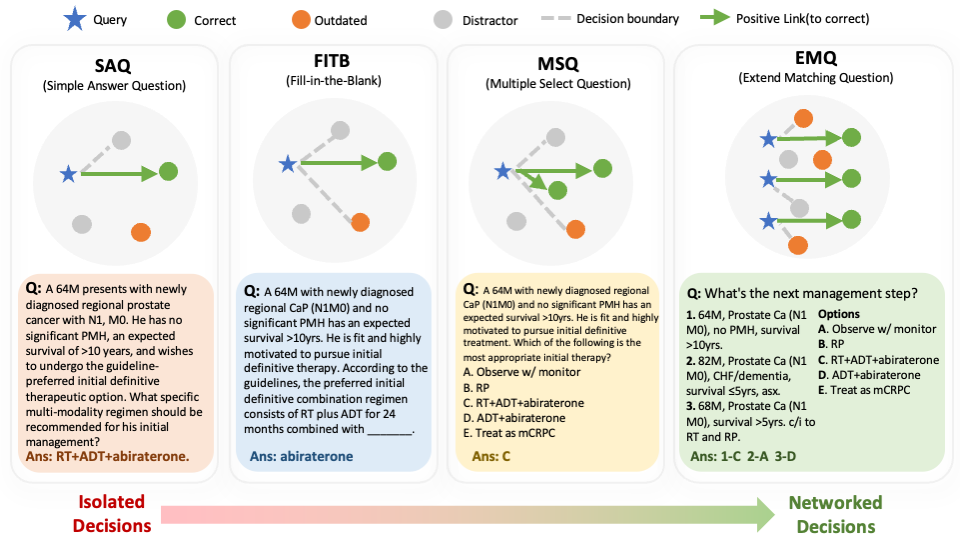}
    \caption{Overview of supervision formats repurposed for SFT. SAQ, FITB, and MSQ rely on isolated, one-to-one or one-to-few decision structures. EMQ introduces a networked structure, requiring the model to resolve many-to-many correspondences within a shared answer pool.
    }
    \label{fig:motivation}
\end{figure*}

In this paper, we bridge this gap by introducing \textbf{SEER-Bench}, a temporally anchored oncology staging benchmark curated from the latest released SEER Research Data~\citep{seer2025}, which contains staging information that strictly aligns with up-to-date NCCN guidelines. The benchmark contains 1{,}992 curated cases across 16 cancer types. We conduct a format-varied study that renders identical NCCN medical update events into four supervision formats~\citep{nccn2025} (Figure~\ref{fig:motivation}), using a shared adaptation setup. Among same-budget SFT variants, \textsc{EMQ} yields the strongest external generalization, reaching 64.8\% SEER-Bench answer accuracy with Qwen3-4B~\citep{qwen2025qwen34b2507} and, in a descriptive rather than controlled comparison, falling within the range of several much larger external reference systems evaluated under the same protocol. Input-density, representation-preservation, and cost-benefit analyses provide diagnostic evidence consistent with this advantage.

Our contributions are as follows:
\begin{itemize}
    \item We introduce SEER-Bench, a real-world oncology staging benchmark with an explicit temporal anchor for evaluating time-sensitive medical reasoning in LLMs.
    \item We show that \textsc{EMQ} provides the strongest external transfer and retention among same-budget SFT variants.
    \item We provide input-density, representation-preservation, and cost-benefit analyses that link \textsc{EMQ}'s advantage to dense clinical contrast signals and more economical representational change.
    \item We release SEER-Bench, related code and prompts at \url{https://github.com/Iflytek-Medical-SouthChina/SeerBench}.
\end{itemize}

\section{Related Work}
\label{sec:related}

\paragraph{Medical QA benchmarks and knowledge temporality.}
Medical QA benchmarks commonly use licensing-exam formats such as MCQs, extended matching items, and clinical vignettes~\citep{jin2021disease,jin2019pubmedqa,pal2022medmcqa,kung2023performance,singhal2025toward,geng2026promedical}. Most, however, evaluate long-stable textbook knowledge rather than temporally versioned guideline transitions. Newer benchmarks improve task realism or contamination control~\citep{arora2025healthbench,yan2026livemedbench,quan2026blind}, but do not directly target clinical update events; temporal QA studies further show that LLMs can rely on stale pretraining facts~\citep{vu2024freshllms,mousavi2024dyknow}. SEER-Bench addresses this gap by evaluating oncology staging from the latest versioned SEER Research Data release~\citep{seer2025}, with versioned registry fields, cohort filtering, and tumor-site coverage.

\paragraph{Mechanisms for updating LLM knowledge.}
LLM knowledge updating has been studied through retrieval-augmented generation, which surfaces external evidence at inference time~\citep{lewis2020retrieval,vu2024freshllms}; continual pretraining, which refreshes parameters at high computational cost~\citep{jang2022temporalwiki}; knowledge editing, which modifies model behavior locally~\citep{meng2022mass,zheng2023can}; and parameter-efficient fine-tuning such as LoRA~\citep{hu2021lora,ge2024time}. These lines mainly differ in how updates are injected. Our study asks a complementary question: under a shared updating setup, how does the supervision format used to present the update affect downstream transfer?

\paragraph{Supervision format in medical training.}
Knowledge-centric medical training data is often rendered as MCQs with distractors~\citep{sileo2024generating}, mixed QA formats~\citep{qiu2025training}, or cloze-style probes~\citep{petroni2019language}. EMQs, a standard medical-exam format~\citep{frey2022item,case1993extended}, require many-to-many matching within a shared candidate space. Prior training-data studies often change content, difficulty, format, and optimization objectives together~\citep{wang2023far,mukherjee2023orca}, making the role of format difficult to inspect. We therefore compare the same update events rendered as \textsc{SAQ}, \textsc{MSQ}, \textsc{FITB}, and \textsc{EMQ} under a same-budget updating protocol.

\section{Same-Budget Updating Protocol}
\label{sec:methods}

\begin{figure*}[t] 
    \centering
    \includegraphics[width=1\linewidth]{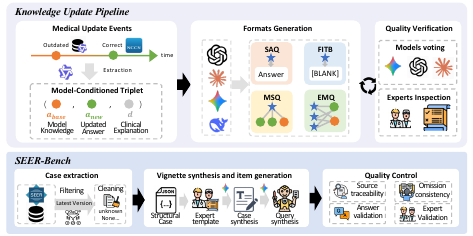}
    \caption{Construction pipeline for update supervision and SEER-Bench. The top panel converts NCCN guideline changes into structured update units and four verified supervision formats. The bottom panel selects SEER Research Data cases, constructs clinical vignettes, and applies multi-stage quality control.}
    \label{fig:pipeline}
\end{figure*}

\subsection{Medical Update Events}
\label{sec:update-events}

The atomic unit of supervision is a \emph{medical update event}, represented as a triplet $(a_{\mathrm{base}}, a_{\mathrm{new}}, d)$. Here, $a_{\mathrm{new}}$ denotes the recommendation supported by the current guideline standard, $a_{\mathrm{base}}$ denotes the alternative recommendation elicited from the target base model when it rejects $a_{\mathrm{new}}$, and $d$ describes the clinically relevant contrast between them. Thus, $a_{\mathrm{base}}$ operationalizes the model's non-current medical knowledge, rather than requiring every alternative to be traceable to a historical guideline version.

As illustrated in the top panel of Figure \ref{fig:pipeline}, in the first stage, model-conditioned triplet construction, we derive current recommendations $a_{\mathrm{new}}$ from NCCN clinical practice guideline in oncology~\citep{nccn2025} and query the target base model to identify non-current medical knowledge. For each $a_{\mathrm{new}}$, the model judges whether the recommendation is correct; when it rejects the current recommendation, we elicit its alternative answer as $a_{\mathrm{base}}$ and ask it to explain the clinical contrast $d$. The resulting triplets therefore target cases where the base model does not reliably encode the current recommendation. The full construction prompt is shown in Figure~\ref{fig:prompt-triplet}.

In the second stage, format rendering, we use Gemini 3.1 Pro~\citep{deepmind2026gemini31pro} with format-specific prompts (Appendix~\ref{sec:appendix-prompts}) to instantiate the model-conditioned triplets as training items. We denote by $\mathcal{R}_f$ the prompt-specified rendering rule for format $f$. For each format $f \in \{\textsc{SAQ}, \textsc{FITB}, \textsc{MSQ}, \textsc{EMQ}\}$, each item is generated as
\begin{equation}
\label{eq:format-rendering-rule}
q_i^{(f)}
=
\mathcal{R}_f\!\left(a_{\mathrm{new},i}, B_i, D_i, s_i\right),
\end{equation}
where $a_{\mathrm{new},i}$ is the unique current recommendation targeted by item $q_i$, $B_i$ is the set of elicited base-model alternatives, $D_i$ contains the corresponding clinical contrasts, and $s_i$ is the source reference material. The prompt requires each item to reflect the contrast between $a_{\mathrm{new},i}$ and $B_i$ so that supervision remains anchored to the intended update. Duplicate $a_{\mathrm{new}}$ targets are removed before rendering; thus each training item targets one current recommendation, although it may contain multiple base-model alternatives.

\subsection{Human-in-the-Loop Quality Verification}
\label{sec:quality-verification}

Quality verification follows an iterative human-in-the-loop process. After each rendering round, we sample a stratified 10\% subset of generated items for expert review. Clinical reviewers assess whether the item correctly reflects the intended update target, whether the clinical vignette is plausible, whether the answer and rationale are consistent with the source recommendation, and whether the distractors or non-current alternatives are clinically meaningful. Reviewer feedback is then incorporated into the rendering process, and the revised prompts are used in the next generation round.

We repeat this review-and-revision cycle until the generated items reach a stable quality level. In the final audit, 94.8\% of reviewed items pass human quality inspection. Items that fail the audit are either revised or removed before training. Details on expert profiles, annotation protocols, and the iterative verification procedure are provided in Appendix~\ref{sec:appendix-quality-verification} and Algorithm~\ref{alg:hitl-quality-verification}.
All formats are constructed from the same pool of 2{,}530 unique vignettes; Appendix~\ref{sec:appendix-vignette-token-controls} provides token-budget and matched-vignette controls.

\subsection{Supervision Formats}
\label{sec:data-construction}

All formats are generated from the same update targets and source references, using format-specific prompts that require clinical case analysis rather than isolated factual recall. Each item is anchored to the contrast between the current recommendation $a_{\mathrm{new}}$, the base-model alternatives $B_i$, and the clinical contrast explanations $D_i$. Thus, all formats expose both the target update and clinically plausible non-current alternatives; the controlled variable is how these contrasts are structured within the training instance. The full rendering prompts are provided in Appendix~\ref{sec:appendix-prompts}.

\begin{itemize}
    \item \textsc{SAQ} (Short Answer Question): presents a clinical case and asks for the current recommendation, the rationale, and the contrast with non-current alternatives.
    \item \textsc{MSQ} (Multiple Select Question): presents a clinical case with item-specific answer options, including clinically plausible alternatives, and asks the model to select all options supported by the current recommendation.
    \item \textsc{FITB} (Fill-in-the-Blank): masks the critical updated medical entity in a clinical narrative and asks the model to complete it, with explanation reflecting the clinical contrast.
    \item \textsc{EMQ} (Extended Matching Question): groups related clinical vignettes into a shared option space and asks the model to match each vignette to the current recommendation.
\end{itemize}

The key structural distinction is therefore not whether a format contains contrastive information, but how that information is organized. In \textsc{SAQ}, \textsc{MSQ}, and \textsc{FITB}, contrast is local to a single clinical vignette: \textsc{SAQ} expresses it in the rationale, \textsc{MSQ} expresses it through item-specific alternatives, and \textsc{FITB} expresses it through the masked entity and explanation. \textsc{EMQ} instead jointly presents multiple related vignettes and candidate answers in a shared option space, requiring cross-vignette discrimination among clinically similar alternatives and creating a many-to-many relational supervision signal.

\subsection{Co-Presented Contrast Structure}
\label{sec:embedding-space-formalization}

We formalize the structural difference among formats by the contrastive alternatives co-presented within a training instance. For item $i$, let $a_{\mathrm{new},i}$ be the current recommendation and let $\mathcal{N}_i^{(f)}$ denote the non-current or clinically plausible alternatives exposed by format $f$ through the vignette, rationale, answer options, or explanation. The co-presented contrast set is
\begin{equation}
\label{eq:co-presented-contrast-set}
\mathcal{Z}_i^{(f)}
=
\{a_{\mathrm{new},i}\}\cup \mathcal{N}_i^{(f)},
\end{equation}
All formats expose contrastive information, but they differ in the scope over which this contrast is organized.

In \textsc{SAQ}, \textsc{FITB}, and \textsc{MSQ}, the contrast set is local to one clinical vignette. \textsc{SAQ} expresses the contrast through the rationale, \textsc{FITB} through the masked updated entity and explanation, and \textsc{MSQ} through item-specific answer options. Although \textsc{MSQ} may contain multiple correct or incorrect options, these alternatives are still organized around a single vignette.

\textsc{EMQ} changes the scope of contrast. For a block $G$ of related update items, it jointly presents multiple vignettes and a shared option space, yielding the block-level contrast set
\begin{equation}
\label{eq:emq-block-contrast-set}
\mathcal{Z}_{G}^{(\textsc{EMQ})}
=
\bigcup_{j\in G}\mathcal{Z}_j^{(\textsc{EMQ})},
\end{equation}
Thus each vignette is interpreted not only against its own local alternatives, but also against alternatives tied to adjacent clinical updates in the same block. This produces a denser relational supervision signal: clinically similar recommendations are co-presented within a single instance, forcing cross-vignette discrimination among updated and non-current alternatives.

This view motivates the diagnostics in Section~\ref{sec:analysis}: if \textsc{EMQ}'s advantage comes from networked contrast rather than merely longer text, it should expose more clinically relevant entities, place related entities closer together, and preserve discriminative representations with less unnecessary movement. A candidate-level gradient interpretation is provided in Appendix~\ref{sec:appendix-format-formalization}.
\section{SEER-Bench}
\label{sec:seer-bench}

Evaluating the format-controlled updating protocol above requires a benchmark with an explicit knowledge temporal anchor: the model must reason under a versioned clinical time boundary rather than merely rely on long-stable textbook knowledge. To this end, we construct \textbf{SEER-Bench} (Figure \ref{fig:pipeline}, bottom panel), which evaluates structured staging reasoning on cases curated from the latest released SEER Research Data~\citep{seer2025}. The benchmark contains 1{,}992 cases across 16 cancer-type groups, and every item undergoes double-blind expert review in alignment with the latest NCCN guideline to verify case fidelity, staging correctness, and rationale validity; disagreements are resolved by adjudication before inclusion. We report two complementary metrics: answer accuracy, the proportion of cases whose predicted staging/classification label is completely correct, and rationale accuracy, the stricter proportion whose final answer and supporting clinical rationale are both correct. The complete construction pipeline, cohort filtering, tumor-site distribution, expert-review protocol, metric definitions, and rationale-grading prompt are detailed in Appendix~\ref{sec:appendix-seer}.

\section{Experiments}
\label{sec:results}

This section evaluates whether the supervision format affects external generalization after medical knowledge updating under a shared adaptation setup and matched training budget, using Qwen3-4B~\citep{qwen2025qwen34b2507} as the primary backbone.
We focus primarily on two newer English-language external evaluations: SEER-Bench tests transfer to temporally anchored oncology staging, whereas HealthBench Professional (hereafter, HealthBench)~\citep{hicks2026healthbench} tests post-update retention on real clinician-chat tasks. MedGUIDE~\citep{li2025medguide} serves strictly as a narrow diagnostic rather than a global benchmark. Its HCC slice remains unchanged over time, serving as a static stability check. Conversely, the NSCLC slice has changed drastically, acting as a knowledge-conflict check where up-to-date models are penalized by outdated ground truth. 
Additional experimental details are provided in Appendix~\ref{sec:appendix-exp-setup}, and the Base RAG inference prompt is in Appendix~\ref{sec:appendix-inference-prompts}; the auxiliary Chinese-language transfer evaluation and representative MedGUIDE cases are in Appendices~\ref{sec:appendix-chinese-eval} and~\ref{sec:appendix-cases}.

\subsection{External Transfer and Retention}
\label{sec:main-results}

\begin{table*}[!t]
\centering
\small
\setlength{\tabcolsep}{4.5pt}
\renewcommand{\arraystretch}{1.12}
\begin{tabularx}{\textwidth}{@{}>{\raggedright\arraybackslash}X
@{\hspace{0.40cm}}C{1.08cm} C{1.55cm}
@{\hspace{0.18cm}}C{1.90cm}
@{\hspace{0.18cm}}C{1.10cm} C{1.25cm}@{\hspace{0.35cm}}}
\toprule
\textbf{System / update} &
\multicolumn{2}{c}{\textbf{SEER-Bench}} &
\textbf{HealthBench} &
\multicolumn{2}{c}{\textbf{MedGUIDE}} \\
\cmidrule(lr){2-3}\cmidrule(lr){4-4}\cmidrule(lr){5-6}
& Acc.~$\uparrow$ & Rat. Acc.~$\uparrow$ & Score~$\uparrow$ & HCC~$\uparrow$ & NSCLC \\
\midrule
\rowcolor{PanelFill}
\multicolumn{6}{@{}l}{\textbf{A. Same-budget Qwen3-4B updates}} \\
Base model & 57.6 & 50.7 & 0.247 & 63.1 & 32.7 \\
SFT, \textsc{EMQ} & \best{64.8} & \best{59.6} & \best{0.263} & \best{82.1} & 23.2 \\
SFT, \textsc{MSQ} & 61.7 & \second{55.1} & 0.243 & \second{72.6} & 24.7 \\
SFT, \textsc{FITB} & 61.5 & 50.9 & 0.237 & 66.4 & 28.7 \\
SFT, \textsc{SAQ} & 62.7 & 52.1 & \second{0.253} & 67.9 & 26.8 \\
RAG, Base RAG & 59.6 & 47.2 & 0.177 & 66.3 & 25.0 \\
RAG, CARE & 60.0 & 52.0 & 0.195 & 69.8 & 20.2 \\
Editing, RECIPE & \second{63.8} & 49.2 & 0.099 & 65.5 & 25.6 \\
Editing, AlphaEdit & 57.9 & 48.0 & 0.087 & 55.7 & 20.8 \\
\midrule
\rowcolor{PanelFill}
\multicolumn{6}{@{}l}{\textbf{B. External systems without task-specific updating}} \\
Qwen3-235B & 59.3 & 52.3 & 0.297 & 78.6 & 47.0 \\
Kimi-2.5 & 60.6 & 51.9 & 0.463 & 66.7 & 25.0 \\
GLM-5 & 63.2 & 55.7 & 0.439 & 77.4 & 58.6 \\
DeepSeek-R1 & 62.3 & 58.5 & 0.377 & 75.0 & 65.8 \\
Claude-Opus-4.6 & \best{67.3} & \best{64.2} & 0.503 & 73.8 & 50.6 \\
Gemini-3.1-Pro & \second{65.6} & 62.0 & \second{0.509} & \best{90.5} & 85.1 \\
GPT-5.4 & 65.1 & \second{63.4} & \best{0.538} & \second{89.3} & 82.1 \\
\medmark{} MediTron-70B & 58.1 & 49.5 & 0.262 & 56.3 & 23.6 \\
\medmark{} HuatuoGPT-O1-70B & 60.2 & 57.4 & 0.354 & 68.5 & 36.0 \\
\bottomrule
\end{tabularx}
\caption[Full English benchmark results.]{
Full English benchmark results. Panel A reports same-budget Qwen3-4B update results across supervision formats. Panel B provides descriptive external-system context without task-specific updating, and \protect\medmark{} marks medical LLMs. Within the MedGUIDE slices, HCC represents the only subset with unchanged NCCN guideline, while NSCLC reflects a largely outdated knowledge base and is excluded from the best/second-best performance rankings. Bold and underline denote the best and second-best values within each panel.}
\label{tab:full-benchmark-results}
\end{table*}

In the controlled SFT rows of Table~\ref{tab:full-benchmark-results}, the same update content yields substantially different external-generalization outcomes when rendered in different supervision formats.
Among the SFT supervision formats, \textsc{EMQ} gives the strongest transfer to temporally anchored oncology staging, improving SEER-Bench answer accuracy from 57.6 to 64.8 and rationale accuracy from 50.7 to 59.6, while modestly increasing the HealthBench score from 0.247 to 0.263.
By contrast, \textsc{MSQ}, \textsc{FITB}, and \textsc{SAQ} yield smaller SEER-Bench gains and less favorable HealthBench retention than \textsc{EMQ}, suggesting that the supervision format influences whether medical updates translate into stable, transferable model behavior.

The retrieval and knowledge-editing baselines reveal a failure mode of medical knowledge updating.
Base RAG and CARE yield only modest SEER-Bench changes while reducing the HealthBench score to 0.177 and 0.195, respectively; RECIPE and AlphaEdit similarly reduce it to 0.099 and 0.087.
These results suggest an unfavorable transfer-retention trade-off: in-context evidence injection and highly localized parameter edits provide limited SEER-Bench transfer while degrading broader clinician-facing medical performance.

The reference-model rows further reveal a mismatch between broad clinician-facing medical capability on HealthBench and temporally anchored oncology performance on SEER-Bench.
The reference LLMs obtain higher HealthBench scores than the controlled Qwen3-4B variants, ranging from 0.297 to 0.538, but their SEER-Bench answer accuracy does not increase monotonically with HealthBench performance.
As descriptive context rather than a controlled model-scale comparison, the reference-model rows provide a scale for interpreting the magnitude of the controlled-format gains: Qwen3-4B updated with \textsc{EMQ} reaches 64.8 SEER-Bench answer accuracy, which falls within the range of several much larger reference systems evaluated under the same prompting protocol.
These results suggest that gains from format-controlled medical updating only partially overlap with broader clinician-facing medical performance.

We also repeat the controlled protocol on Llama-3.1-8B-Instruct~\citep{grattafiori2024llama} as an additional backbone check (Appendix~\ref{sec:appendix-llama-results}). The same qualitative pattern holds: \textsc{EMQ} is again the strongest SFT format on SEER-Bench, improving answer accuracy from 58.0 to 65.2 and rationale accuracy from 51.1 to 60.0, while slightly increasing HealthBench from 0.260 to 0.276. MedGUIDE again shows a boundary condition, with \textsc{FITB} performing best on the NSCLC slice. Statistical tests for the main controlled comparisons are reported in Appendix~\ref{sec:appendix-significance}.

\subsection{Older Direct-Acquisition Benchmarks Can Be Misleading}
\label{sec:medguide-discussion}

\begin{table}[t]
\centering
\small
\renewcommand{\arraystretch}{1.12}
\setlength{\tabcolsep}{7.24pt}
\begin{tabular}{lcccc}
\toprule
\textbf{Format} &
\multicolumn{2}{c}{\textbf{SEER-Bench}} &
\multicolumn{2}{c}{\textbf{MedGUIDE}} \\
\cmidrule(lr){2-3}\cmidrule(lr){4-5}
 & HCC~$\uparrow$ & NSCLC~$\uparrow$ & HCC~$\uparrow$ & NSCLC \\
\midrule
Base & 63.9 & 44.8 & 63.1 & 32.7 \\
\textsc{EMQ}  & \best{80.8} & \best{58.4} & \best{82.1} & 23.2 \\
\textsc{MSQ}  & \second{78.6} & \second{56.5} & \second{72.6} & 24.7 \\
\textsc{FITB} & 72.3 & 55.1 & 66.4 & 28.7 \\
\textsc{SAQ}  & 74.2 & 53.2 & 67.9 & 26.8 \\
\bottomrule
\end{tabular}
\caption{Cancer-type slice accuracy on SEER-Bench and MedGUIDE. The SEER-Bench columns and the MedGUIDE HCC column are keyed to current NCCN standards, whereas the MedGUIDE NSCLC answer key is largely outdated and is excluded from the ranking.}
\label{tab:cancer-slice}
\end{table}

While SEER-Bench and HealthBench characterize external transfer and retention, MedGUIDE provides a view of direct acquisition on earlier update slices. Despite being published during the same historical period, its expert-selected HCC and NSCLC slices differ in their temporal stability: after expert-verification, HCC slice remains mostly unchanged between the historical and latest versions, whereas the NSCLC slice changes drastically. We therefore treat MedGUIDE not as a global ranking of supervision formats, but as a diagnostic tool for understanding how direct-acquisition scores interact with the volatility and age of specific benchmark slices.

The two slices show different patterns. On HCC, \textsc{EMQ} achieves the highest score, consistent with its stronger transfer behavior on SEER-Bench. On NSCLC, however, \textsc{FITB} obtains the highest score. Rather than indicating a general advantage of \textsc{FITB}, this result suggests that masked-span supervision may leave the model more compatible with older, locally memorized knowledge. In other words, a high score on an older slice can partly reflect persistence of non-current knowledge rather than successful updating toward the current standard.

Table~\ref{tab:cancer-slice} quantifies this contrast and extends the same cancer-type slices to SEER-Bench. An expert disease-level audit finds that the NCCN guideline for HCC updates about 1--3 times per year, and 92.2\% of MedGUIDE HCC questions remain concordant with the latest NCCN version. While NSCLC guideline updates 4--6 versions per year and only 43.8\% of questions remaining concordant. On the three slices keyed to current knowledge (SEER-Bench HCC/NSCLC and MedGUIDE HCC), \textsc{EMQ} leads consistently, whereas on the outdated-keyed MedGUIDE NSCLC slice every updated model scores below the base model. Such persistence of stale associations is precisely what makes outdated knowledge clinically hazardous: a staging error can change treatment intensity, and a stale recommendation or an overconfident rationale can delay or misdirect appropriate care (Appendix~\ref{sec:appendix-clinical-safety} presents concrete guideline-versioned cases).

Thus, the mixed MedGUIDE results highlight a limitation of older direct-acquisition benchmarks: they can reward formats that preserve or recover stale local associations. MedGUIDE therefore complements the main results by showing why external, temporally anchored evaluations such as SEER-Bench are necessary for assessing whether medical updates generalize beyond direct slice acquisition.

\subsection{Effect of Knowledge-Update Frequency}
\label{sec:update-frequency}

\begin{table}[t]
\centering
\small
\renewcommand{\arraystretch}{1.12}
\setlength{\tabcolsep}{4.07pt}
\begin{tabular}{lcccc}
\toprule
\textbf{Format} & \textbf{Overall} &
\textbf{Reversion} & \textbf{Frequent} & \textbf{Lower-upd.} \\
 & & (2.5\%) & (59.9\%) & (37.6\%) \\
\midrule
Base & 57.6 & 56.0 & 54.5 & 62.6 \\
\textsc{EMQ}  & \best{64.8} & \best{62.0} & \best{62.4} & \best{68.9} \\
\textsc{MSQ}  & 61.7 & \best{62.0} & \second{60.9} & 63.2 \\
\textsc{FITB} & 61.5 & 54.0 & 58.9 & 66.1 \\
\textsc{SAQ}  & \second{62.7} & 56.0 & 60.1 & \second{67.3} \\
\bottomrule
\end{tabular}
\caption{SEER-Bench answer accuracy by temporal-update stratum. \emph{Reversion}: a return to an earlier staging rule would change the current case's stage. \emph{Frequent}: the applicable staging rules changed in at least three successive revisions. \emph{Lower-upd.}: all remaining cases that changed in fewer revisions.}
\label{tab:update-frequency}
\end{table}

A natural question for a temporally anchored benchmark is whether the observed format effects vary with how frequently the underlying knowledge changes. We therefore conduct an expert-verified post-hoc stratification of all 1{,}992 SEER-Bench cases into three mutually exclusive temporal-update strata (Table~\ref{tab:update-frequency}). To keep this audit independent of the LLM-generated alternatives, the strata are assigned from the versioned revision history of the applicable staging rules rather than from the generated candidates.

\textsc{EMQ} achieves the highest or tied-highest point estimate in every stratum. Frequent-update cases are substantially harder before adaptation (base accuracy 54.5\% against 62.6\% on lower-update-density cases), yet the adaptation gains are not attenuated there and remain strongly format-dependent, with \textsc{EMQ} strongest on this stratum covering 59.9\% of the benchmark. On reversion-sensitive cases, \textsc{EMQ} and \textsc{MSQ} both reach 62.0\%, against 56.0\% for both base model and \textsc{SAQ}, and 54.0\% for \textsc{FITB}, consistent with explicit candidate sets helping distinguish competing current and non-current associations; \textsc{MSQ}, however, gains little on lower-update-density cases, whereas \textsc{EMQ} remains effective across all strata, suggesting that its shared candidate pool induces more transferable decision boundaries than discrimination tied to a single local option set.
\section{Understanding the EMQ Advantage}
\label{sec:analysis}

Section~\ref{sec:results} shows that \textsc{EMQ} provides the strongest external transfer and retention among the same-budget SFT variants. We examine this pattern through a diagnostic input-to-representation chain: \textsc{EMQ} exposes denser clinical contrast signals while preserving discriminative representations with less movement from the base model. Implementation details for the representation probes are provided in Appendix~\ref{sec:appendix-impl}.

\subsection{Clinical Relation Signal Density}
\label{sec:mechanism-summary}

\begin{table*}[t]
\centering
\scriptsize
\renewcommand{\arraystretch}{1.08}
\setlength{\tabcolsep}{3.71pt}
\begin{tabular}{lrrrrrrrrrrr}
\toprule
\textbf{Format} &
\multicolumn{6}{c}{\textbf{Training supervision}} &
\multicolumn{5}{c}{\textbf{SEER-Bench inference}} \\
\cmidrule(lr){2-7}\cmidrule(lr){8-12}
 & \textbf{Q tok.} & \textbf{A tok.} &
\textbf{Q ent.}~$\uparrow$ & \textbf{A ent.}~$\uparrow$ &
\textbf{Step share}~$\uparrow$ & \textbf{Ent. dist.}~$\downarrow$ &
\textbf{A ent.}~$\uparrow$ & \textbf{Step share}~$\uparrow$ &
\textbf{Ent. dist.}~$\downarrow$ & \textbf{1024-tok Acc.}~$\uparrow$ &
\textbf{1024-tok Rat.}~$\uparrow$ \\
\midrule
Base & -- & -- & -- & -- & -- & -- & 22.2 & 1.27\% & 60.12 & 56.5 & 49.8 \\
\midrule
\textsc{EMQ}  & 258.7 & 398.4 & \textbf{11.4} & \textbf{12.6} & \textbf{2.03\%} & \textbf{52.58} & \textbf{29.6} & \textbf{3.44\%} & \textbf{31.04} & \textbf{62.9} & \textbf{57.8} \\
\textsc{MSQ}  & 260.2 & 393.9 & 10.7 & 10.6 & 1.07\% & 61.71 & 28.2 & 2.36\% & 38.71 & 59.8 & 53.7 \\
\textsc{FITB} & 254.3 & 388.8 & 5.8 & 8.9 & 1.51\% & 72.26 & 25.2 & 1.49\% & 58.06 & 58.4 & 49.5 \\
\textsc{SAQ}  & 261.5 & 401.1 & 8.9 & 11.3 & 1.27\% & 58.64 & 27.3 & 2.69\% & 34.78 & 61.2 & 51.3 \\
\bottomrule
\end{tabular}
\caption{Diagnostic clinical-relation signal density in training supervision and SEER-Bench inference. Q tok.\ and A tok.\ denote average prompt and answer length; Q ent.\ and A ent.\ denote the average number of cancer-related entities per item; Step share is the percentage of period-delimited reasoning steps containing cancer-related answer entities; Ent.\ dist.\ is the average token distance between adjacent cancer-related entities, where lower indicates denser local relations. The 1024-token columns are computed in a separate SEER-Bench run with the same generation-token limit and therefore may differ from Table~\ref{tab:full-benchmark-results}.}
\label{tab:signal-density}
\end{table*}

Table~\ref{tab:signal-density} provides a diagnostic check of whether \textsc{EMQ}'s structural differences are visible before representation-level analysis. The four training corpora have comparable prompt and answer lengths, but \textsc{EMQ} exposes the most tumor-related entities in both prompts and answers, has the largest share of reasoning steps containing tumor-related answer entities, and yields the shortest average distance between related entities. This pattern is consistent with the view that \textsc{EMQ}'s shared option pool does not merely add text, but organizes isolated update facts into a denser local clinical-relation structure.

A similar signal-density pattern appears in SEER-Bench outputs after updating. Compared with the base model, all format-updated variants generally increase the number of tumor-related entities in the response, but \textsc{EMQ} produces the largest increase and the shortest distance between related entities. With the same 1024-token generation limit, \textsc{EMQ} still achieves the best answer and rationale accuracy. These diagnostics make a pure length-based explanation less likely: \textsc{EMQ} is associated with responses that concentrate a fixed reasoning budget into a denser structure of mutually constraining oncology entities. Appendix~\ref{sec:appendix-density-network-ablation} further probes this potential confound with density-matched \textsc{MSQ} and shuffled-\textsc{EMQ} controls. To further rule out the possibility that \textsc{EMQ}'s gains are driven merely by more unique vignettes or a larger token budget, Appendix~\ref{sec:appendix-vignette-token-controls} reports vignette/token accounting and two matched-vignette controls. All formats are rendered from the same pool of 2{,}530 unique vignettes, and \textsc{EMQ} remains strongest when vignette exposure is matched or when single-vignette \textsc{EMQ} instances are constructed.

\subsection{Representation Economy}
\label{sec:analysis-l2}

Table~\ref{tab:repr-summary} connects the input- and output-side patterns above to representation-level diagnostics. \textsc{EMQ} has the smallest mean L2 displacement across evaluated layers, the smallest final-layer L2 displacement, and the highest mean CKA, while retaining tied-best final-layer linear-probe accuracy. This pattern suggests that denser clinical relation signals are associated with transfer gains without broad representational displacement; instead, \textsc{EMQ} preserves more of the base model's class-discriminative structure as measured by the probe.

\begin{table}[t]
\centering
\small
\renewcommand{\arraystretch}{1.12}
\setlength{\tabcolsep}{4.22pt}
\begin{tabular}{lrrrr}
\toprule
\textbf{Format} &
\textbf{L2 avg.}~$\downarrow$ &
\textbf{L2 final}~$\downarrow$ &
\textbf{CKA avg.}~$\uparrow$ &
\textbf{Probe}~$\uparrow$ \\
\midrule
\textsc{EMQ}  & \textbf{0.039} & \textbf{0.097} & \textbf{0.9943} & \textbf{0.9848} \\
\textsc{MSQ}  & 0.049 & 0.111 & 0.9914 & \textbf{0.9848} \\
\textsc{FITB} & 0.047 & 0.114 & 0.9932 & 0.9834 \\
\textsc{SAQ}  & 0.055 & 0.127 & 0.9902 & 0.9779 \\
\bottomrule
\end{tabular}
\caption{Representation-level mechanism summary. L2 avg.\ is averaged over all evaluated layers; L2 final and Probe are computed at the final layer.}
\label{tab:repr-summary}
\end{table}

\begin{figure}[h]
    \centering
    \includegraphics[width=\columnwidth]{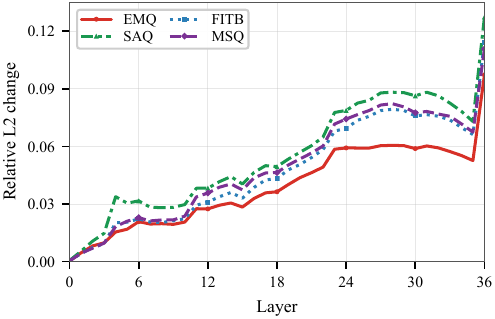}
    \caption{Layer-wise L2 ratio relative to the Qwen3-4B. Lower indicates a smaller representational change.}
    \label{fig:l2-ratio}
\end{figure}

Figure~\ref{fig:l2-ratio} shows that this representational preservation holds across layers and is most visible in the deep layers. \textsc{EMQ}'s L2 ratio is 0.058 over layers 28--35 versus 0.084 for \textsc{SAQ}, and 0.097 at the final layer versus 0.127 for \textsc{SAQ}, a 24\% reduction. This pattern is consistent with the view that dense matching supervision guides the update toward task-relevant directions while avoiding broad representational drift.

The clustering diagnostics in Appendix~\ref{sec:appendix-analysis-tables} reveal a useful ranking conflict: unsupervised compactness metrics favor \textsc{FITB} or \textsc{SAQ}, whereas the linear probe favors \textsc{EMQ} and \textsc{MSQ}. This suggests that tighter unsupervised clusters are not necessarily better for knowledge updating; excessive compression of intra-class variance may remove fine-grained clinical substructure needed for downstream discrimination.

\subsection{Cross-Scale Cost-Benefit View}
\label{sec:analysis-goldilocks}

\begin{figure}[t]
    \centering
    \includegraphics[width=\columnwidth]{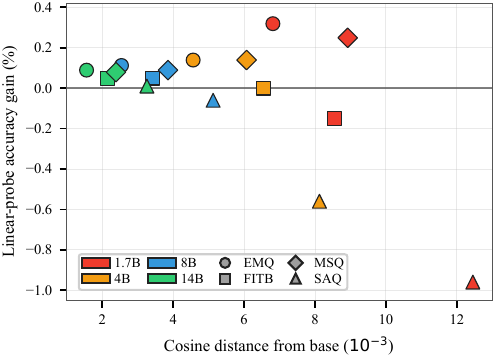}
    \caption{Final-layer cost-benefit comparison. $X$-axis is cosine distance from the base model; $Y$-axis is the change in linear-probe accuracy over the base model.}
    \label{fig:goldilocks}
\end{figure}

Figure~\ref{fig:goldilocks} extends the final-layer cost-benefit comparison across 1.7B, 4B, 8B, and 14B Qwen3 backbones~\citep{yang2025qwen3}. Across all evaluated scales, \textsc{EMQ} lies in the low-movement, positive-gain region, whereas \textsc{SAQ} more often occupies a higher-movement, lower-gain region, especially for smaller backbones. \textsc{MSQ} can also yield positive probe gains, but generally requires larger representational movement than \textsc{EMQ}.

Appendix~\ref{sec:appendix-cross-scale-seer} reports the corresponding SEER-Bench task results. The same qualitative pattern holds at the task level: \textsc{EMQ} gives the best answer and rationale accuracy for every evaluated backbone, improving answer accuracy from 54.2 to 60.5 on Qwen3-1.7B and from 58.7 to 67.8 on Qwen3-14B. These cross-scale results suggest that the \textsc{EMQ} advantage is not an artifact of the 4B main setting, but a consistent supervision-format pattern across model capacities.
\section{Conclusion}
We studied an often-overlooked variable in medical knowledge updating: whether the supervision format affects generalization under a shared adaptation setup and matched training budget. We introduced SEER-Bench as a temporally anchored oncology-staging benchmark and rendered medical update events into four supervision formats. Across same-budget SFT comparisons, \textsc{EMQ} gives the most stable external transfer and retention, producing competitive SEER-Bench results with an updated 4B model. Our diagnostics suggest that \textsc{EMQ}'s advantage is associated with denser clinical contrast signals and more economical representational movement. Overall, medical knowledge updating should not be treated only as a choice of update algorithm; the structure used to present knowledge as supervision also shapes whether the update transfers to external tasks.

\section*{Limitations}
This study has several limitations. First, SEER-Bench and the update
items rely partly on LLM-assisted construction; although we apply source
tracing, answer-space validation, and expert spot checks, residual
clinical or linguistic errors may remain. Second, the experiments focus
on oncology and mainly on a 4B base model with LoRA-based adaptation, so
the observed format effects may differ across medical domains, model
scales, and update algorithms. Third, although we hold update content and
training budget fixed, the formats differ in length, candidate-set size,
and surface complexity, which may partially contribute to the observed
gaps. Finally, our representation analyses are diagnostic rather than
causal, and the NSCLC slice suggests that EMQ is not uniformly optimal:
slot-like molecular updates may benefit more from the focused
autoregressive signal provided by FITB.
\section*{Ethical Considerations and Data Statement}
This work studies medical knowledge updating as an offline evaluation and model-adaptation problem, not as a deployed clinical decision-support system. The reported models should not be used to make diagnosis, staging, treatment, or medication decisions without qualified clinical oversight.

SEER-Bench is constructed from de-identified, versioned SEER records and is intended solely for developing and evaluating questions related to tumor staging using real-world oncology cases. To reduce construction errors, we employ multiple quality-control procedures, including source tracing, answer-space validation, model-based consensus voting, and expert review. Human involvement was limited to professional annotation and quality verification of de-identified research materials. Complete annotation guidelines, risk disclaimers (explicitly staging minimal risk limited to professional time commitment) and confidentiality agreements are also provided in the annotation process. Reviewers with clinical or biomedical expertise were compensated for their annotation time, as documented in our annotation records. The annotation task did not involve patient contact, intervention, or real-time clinical decision-making, and reviewer-identifying information will not be released.

We integrate references to the National Comprehensive Cancer Network (NCCN) Guidelines® and the Chinese Guideline/ Expert Consensus. These external knowledge sources are utilized under open-use principles solely for academic research, benchmarking, and pedagogical evaluation.  

We publicly release SEER-Bench, and the construction and evaluation prompts and code of the four format supervision at \url{https://github.com/Iflytek-Medical-SouthChina/SeerBench}, subject to source-resource licenses and data-use restrictions. Provenance and use conditions are documented in the accompanying data card.

We also acknowledge the use of Gemini-3.1-Pro for linguistic refinement and editorial suggestions during the manuscript revision.
\section*{Acknowledgments}

This work was supported by the Noncommunicable Chronic Diseases-National Science and Technology Major Project (Grant No.~2023ZD0509100).

\bibliography{custom}

\appendix

\newpage
\section{Data Construction Prompts}
\label{sec:appendix-prompts}

The data construction follows a two-stage prompting pipeline. In the first stage (Figure~\ref{fig:prompt-triplet}), we use the current authoritative guideline text together with the target base model's judgment of the current recommendation to construct structured triplets: the guideline-supported current recommendation, the non-current alternative produced by the base model when it rejects that recommendation, and a clinical explanation of the contrast. In the second stage, each extracted triplet and its source reference material are rendered by Gemini 3.1 Pro into training items in four supervision formats (\textsc{SAQ}, \textsc{MSQ}, \textsc{FITB}, and \textsc{EMQ}; Figures~\ref{fig:prompt-saq},~\ref{fig:prompt-msq},~\ref{fig:prompt-fitb}, and~\ref{fig:prompt-emq}). All format-rendering prompts require clinical case analysis rather than pure factual recall and dual-perspective answering, with both the current-standard answer and the base-model alternative. After generation, candidate items undergo model voting and expert inspection; the manual audit covers a stratified random 10\% sample, and item-construction quality labels achieve Cohen's $\kappa=0.85$.

\begin{figure*}[t]
\centering
\begin{tcolorbox}[
    colback=gray!3!white,
    colframe=gray!70!black,
    title=\textbf{Knowledge Triplet Extraction Prompt},
    fonttitle=\bfseries\large,
    fontupper=\small,
    rounded corners,
    boxrule=0.8pt,
    arc=3mm,
    boxsep=3mm,
    toptitle=2mm,
    bottomtitle=2mm,
    before upper={\renewcommand\baselinestretch{1.1}\parskip=0.3em}
]
\textbf{Role:} A senior oncologist and leading expert in multi-site cancer diagnosis, staging, and evidence-based treatment strategies. 

\textbf{Objective:} Analyze the provided Reference Material to identify oncology-related medical inaccuracies or suspicious claims. Generate question-building knowledge where the distractor is based on the incorrect text, but the correct answer reflects the current authoritative clinical standard.

\textbf{Reference Material:} \texttt{\{guide\}}

\textbf{Negative Constraints:}
\begin{enumerate}[leftmargin=1.2em, topsep=0pt, itemsep=0pt]
    \item No Meta-References: never use phrases such as ``According to the text'', ``The provided material states'', or ``Based on the reference''.
    \item No Rote Memorization of Statistics: do not ask for specific percentages, p-values, hazard ratios, or study-specific response rates.
    \item No Study-Specific Hooks: do not use a clinical-trial name as the primary premise; convert trial findings into generalized clinical decision-making or standard of care.
    \item No Non-Medical Content: ignore formatting, dates, and page numbers; focus on diagnostic criteria, staging, and therapeutic interventions.
\end{enumerate}

\textbf{Knowledge Extraction Principles:}
\begin{enumerate}[leftmargin=1.2em, topsep=0pt, itemsep=0pt]
    \item Clinical Standalone: the extracted medical knowledge must be understandable by a specialist without seeing the reference material.
    \item Knowledge Triplet Mechanism: identify the current guideline-supported recommendation and the base model's non-current alternative, then output a triplet. \texttt{new\_knowledge} must state the current recommendation supported by the reference; \texttt{old\_knowledge} must faithfully represent the base-model alternative rather than a claim from the reference material; \texttt{difference} must explain the clinically meaningful contrast between them.
    \item Extraction Focus (Concept over Tone): extract only concrete factual shifts, such as drug indications, staging criteria, biomarker testing, or specific management steps. Do not extract differences based only on subjective degree, wording preferences, or modal verbs such as ``should'' versus ``must''.
\end{enumerate}

\textbf{Output:} Return only a strictly valid JSON array. Each object contains \texttt{id} and \texttt{knowledge\_triplet}, where \texttt{knowledge\_triplet} contains \texttt{new\_knowledge}, \texttt{old\_knowledge}, and \texttt{difference}.
\end{tcolorbox}
\caption{Prompt template for knowledge triplet extraction (Stage~1). Given authoritative guideline text, Gemini 3.1 Pro identifies outdated or erroneous claims and outputs structured triplets used to construct clinically grounded distractors and updated answers.}
\label{fig:prompt-triplet}
\end{figure*}

The first-stage prompt turns guideline passages into old/new/difference triplets, which are the atomic update units used by the later format renderers. The negative constraints avoid superficial artifacts such as trial-name memorization or meta-references to the source text.

\begin{figure*}[t]
\centering
\begin{tcolorbox}[
    colback=gray!3!white,
    colframe=gray!70!black,
    title=\textbf{SAQ Generation Prompt},
    fonttitle=\bfseries\large,
    fontupper=\small,
    rounded corners,
    boxrule=0.8pt,
    arc=3mm,
    boxsep=3mm,
    toptitle=2mm,
    bottomtitle=2mm,
    before upper={\renewcommand\baselinestretch{1.1}\parskip=0.3em}
]
\textbf{Role:} Senior Oncology Medical Educator designing board-style Short Answer Questions (SAQs).

\textbf{Objective:} Design open-ended SAQs based on the provided Knowledge Triplet and Reference Material. Provide the ideal answer based strictly on the Reference Material (which represents the updated and current standard, i.e., New Knowledge), AND the answer based on the outdated standard (Old Knowledge).

\textbf{Negative Constraints:}
\begin{enumerate}[leftmargin=1.2em, topsep=0pt, itemsep=0pt]
    \item No Meta-References (do not ask ``What did the guideline change?'')
    \item No Conceptual Repetition
    \item No Pure Factual Recall---must be a clinical case analysis
\end{enumerate}

\textbf{Design Principles:}
\begin{enumerate}[leftmargin=1.2em, topsep=0pt, itemsep=0pt]
    \item Clinical Case Analysis (Mandatory): present a detailed patient vignette. The question stem MUST incorporate clinical contexts of both the outdated standard (Old Knowledge) and the updated standard (New Knowledge). Furthermore, the core of the question must explicitly revolve around the pivotal clinical changes between the old and new knowledge.
    \item Depth of Knowledge: ask for recommended management and rationale
    \item Dual-Perspective Answering: Reference Material standard (New Knowledge) vs.\ outdated standard (Old Knowledge)
\end{enumerate}

\textbf{Output:} JSON with fields: question, new\_knowledge\_answer, new\_knowledge\_rationale, old\_knowledge\_answer, old\_knowledge\_rationale, construction\_rationale, question\_keywords.

\end{tcolorbox}
\caption{Prompt template for SAQ (Short Answer Question) generation.}
\label{fig:prompt-saq}
\end{figure*}

\textsc{SAQ} provides direct open-ended supervision for one clinical shift at a time. Its strength is concise answer-and-rationale training, but it does not force the model to compare several nearby old/new decision boundaries within the same item.

\begin{figure*}[t]
\centering
\begin{tcolorbox}[
    colback=gray!3!white,
    colframe=gray!70!black,
    title=\textbf{MSQ Generation Prompt},
    fonttitle=\bfseries\large,
    fontupper=\small,
    rounded corners,
    boxrule=0.8pt,
    arc=3mm,
    boxsep=3mm,
    toptitle=2mm,
    bottomtitle=2mm,
    before upper={\renewcommand\baselinestretch{1.1}\parskip=0.3em}
]
\textbf{Role:} Senior Oncology Medical Educator designing board-style Multiple Select Questions (MSQs).

\textbf{Objective:} Design clinical MSQs where options explicitly juxtapose the outdated standard (Old Knowledge) against the updated and current standard (New Knowledge). At least 5 options (A--F), with 1--4 correct answers. Provide correct choices based strictly on the Reference Material (which represents the updated and current standard, i.e., New Knowledge) AND choices based on the outdated standard (Old Knowledge).

\textbf{Negative Constraints:}
\begin{enumerate}[leftmargin=1.2em, topsep=0pt, itemsep=0pt]
    \item No Meta-References
    \item No Conceptual Repetition across questions
    \item No Subjective Tone Testing (``should'' vs.\ ``must'')
    \item No Pure Factual Recall---must be a clinical case analysis
\end{enumerate}

\textbf{Design Principles:}
\begin{enumerate}[leftmargin=1.2em, topsep=0pt, itemsep=0pt]
    \item Clinical Case Analysis (Mandatory): detailed patient presentation leading to a management dilemma. The question stem MUST incorporate clinical contexts of both the outdated standard (Old Knowledge) and the updated standard (New Knowledge). Furthermore, the core of the question must explicitly revolve around the pivotal clinical changes between the old and new knowledge.
    \item Old/New Integration (Mandatory): options include practices based on the Reference Material (New Knowledge) AND practices based on the outdated standard (Old Knowledge).
    \item Multiple Select: 5+ options, 1--4 correct answers
    \item Dual-Perspective Answering: Reference Material standard (New Knowledge) vs.\ outdated standard (Old Knowledge)
\end{enumerate}

\textbf{Output:} JSON with fields: scenario, options, new\_knowledge\_answers, new\_knowledge\_rationale, old\_knowledge\_answers, old\_knowledge\_rationale, construction\_rationale, question\_keywords.
\end{tcolorbox}
\caption{Prompt template for MSQ (Multiple Select Question) generation.}
\label{fig:prompt-msq}
\end{figure*}

\textsc{MSQ} increases answer-space complexity by placing old-knowledge and new-knowledge options in the same choice set. This makes the supervision more contrastive than \textsc{SAQ}, although each instance still centers on a single scenario.

\begin{figure*}[t]
\centering
\begin{tcolorbox}[
    colback=gray!3!white,
    colframe=gray!70!black,
    title=\textbf{FITB Generation Prompt},
    fonttitle=\bfseries\large,
    fontupper=\small,
    rounded corners,
    boxrule=0.8pt,
    arc=3mm,
    boxsep=3mm,
    toptitle=2mm,
    bottomtitle=2mm,
    before upper={\renewcommand\baselinestretch{1.1}\parskip=0.3em}
]
\textbf{Role:} Senior Oncology Medical Educator designing Fill-in-the-Blank (FITB) clinical questions.

\textbf{Objective:} Design clinical FITB questions where the blank precisely targets the core conceptual shift. Provide the correct fill based strictly on the Reference Material (which represents the updated and current standard, i.e., New Knowledge) AND the fill based on the outdated standard (Old Knowledge).

\textbf{Negative Constraints:}
\begin{enumerate}[leftmargin=1.2em, topsep=0pt, itemsep=0pt]
    \item No Meta-References
    \item No Conceptual Repetition
    \item No Grammar-based Blanks---must be a crucial medical entity or threshold
    \item No Pure Factual Recall---must describe a specific clinical case scenario
\end{enumerate}

\textbf{Design Principles:}
\begin{enumerate}[leftmargin=1.2em, topsep=0pt, itemsep=0pt]
    \item Clinical Case Analysis (Mandatory): concise patient vignette with demographics, presentation, prior therapies. The clinical statement MUST incorporate clinical contexts of both the outdated standard (Old Knowledge) and the updated standard (New Knowledge). Furthermore, the core of the question (and the targeted blank) must explicitly revolve around the pivotal clinical changes between the old and new knowledge.
    \item Targeted Blanks: case concludes with a management decision where the key entity is [BLANK]
    \item Dual-Perspective Answering: Reference Material standard (New Knowledge) vs.\ outdated standard (Old Knowledge)
\end{enumerate}

\textbf{Output:} JSON with fields: clinical\_statement, new\_knowledge\_fill, new\_knowledge\_rationale, old\_knowledge\_fill, old\_knowledge\_rationale, construction\_rationale, question\_keywords.
\end{tcolorbox}
\caption{Prompt template for FITB (Fill-in-the-Blank) generation.}
\label{fig:prompt-fitb}
\end{figure*}

\textsc{FITB} targets a specific entity, threshold, or management decision. This makes the update signal highly localized, which is useful for slot-like facts but may provide less context for broader clinical reasoning.

\begin{figure*}[t]
\centering
\begin{tcolorbox}[
    colback=gray!3!white,
    colframe=gray!70!black,
    title=\textbf{EMQ Generation Prompt},
    fonttitle=\bfseries\large,
    fontupper=\small,
    rounded corners,
    boxrule=0.8pt,
    arc=3mm,
    boxsep=3mm,
    toptitle=2mm,
    bottomtitle=2mm,
    before upper={\renewcommand\baselinestretch{1.1}\parskip=0.3em}
]
\textbf{Role:} Senior Oncology Medical Educator designing board-style Extended Matching Questions (EMQs).

\textbf{Objective:} Design clinical EMQs where the shared options list contains elements from BOTH the outdated standard (Old Knowledge) and the updated standard (New Knowledge). Provide the correct answer based strictly on the Reference Material (which represents the updated and current standard, i.e., New Knowledge) AND the answer based on the outdated standard (Old Knowledge), for each vignette.

\textbf{Negative Constraints:}
\begin{enumerate}[leftmargin=1.2em, topsep=0pt, itemsep=0pt]
    \item No Meta-References
    \item No Text-Matching (no wordplay or terminology formatting tests)
    \item No Conceptual Repetition across vignettes
    \item No Pure Factual Recall---must be clinical case analyses
\end{enumerate}

\textbf{Design Principles:}
\begin{enumerate}[leftmargin=1.2em, topsep=0pt, itemsep=0pt]
    \item EMQ Structure: single Theme, shared Options (A--H), $\geq$2 distinct Clinical Vignettes
    \item Clinical Case Analysis (Mandatory): each vignette is a patient case requiring a clinical decision. Each vignette MUST incorporate clinical contexts of both the outdated standard (Old Knowledge) and the updated standard (New Knowledge). Furthermore, the core of the vignettes must explicitly revolve around the pivotal clinical changes between the old and new knowledge.
    \item Old/New Integration (Mandatory): options include practices based on the Reference Material (New Knowledge) AND practices based on the outdated standard (Old Knowledge).
    \item Dual-Perspective Answering per vignette: Reference Material standard (New Knowledge) vs.\ outdated standard (Old Knowledge)
\end{enumerate}

\textbf{Output:} JSON with fields: theme, options, vignettes (each with scenario, new\_knowledge\_answer, new\_knowledge\_rationale, old\_knowledge\_answer, old\_knowledge\_rationale, construction\_rationale, question\_keywords).

\end{tcolorbox}
\caption{Prompt template for EMQ (Extended Matching Question) generation.}
\label{fig:prompt-emq}
\end{figure*}

\textsc{EMQ} groups multiple related vignettes under a shared option pool. The resulting supervision asks the model to distinguish which cases truly changed and which cases should remain stable, matching the stability--plasticity issue analyzed below.
\section{Expert Profile and Annotation Protocols}
\label{sec:appendix-quality-verification}

We used an iterative human-in-the-loop process to verify the generated update-supervision items. After each rendering round, a stratified 10\% subset of generated items was reviewed by clinical experts. Reviewer feedback was incorporated into the prompt and rendering process before the next generation round.

The expert review team consisted of 10 reviewers with clinical or biomedical expertise. Reviewers were compensated at approximately \$2 USD per annotated item. In the final audit, 94.8\% of reviewed items passed human quality inspection. Inter-annotator agreement for binary item-quality labels was Cohen's $\kappa=0.85$. Items that failed the audit were revised or removed before training.

\begin{algorithm}[t]
\small
\caption{Human-in-the-loop quality verification for update-supervision items.}
\label{alg:hitl-quality-verification}
\begin{algorithmic}[1]
\Require Update triplets $\mathcal{U}$; format renderers $\{R_f\}_{f \in \{\textsc{SAQ},\textsc{MSQ},\textsc{FITB},\textsc{EMQ}\}}$; source references $\mathcal{S}$; sampling rate $\rho=0.10$
\Ensure Verified training corpora $\{\mathcal{D}_f\}$

\State Initialize format-specific rendering prompts and constraints.
\Repeat
    \ForAll{formats $f \in \{\textsc{SAQ},\textsc{MSQ},\textsc{FITB},\textsc{EMQ}\}$}
        \State Render candidate items $\tilde{\mathcal{D}}_f = R_f(\mathcal{U}, \mathcal{S})$.
        \State Apply automatic checks for schema validity, duplicate targets, source traceability, answer-space validity, and old/new consistency.
        \State Revise or discard items that fail deterministic checks.
        \State Draw a stratified sample $\mathcal{A}_f \subset \tilde{\mathcal{D}}_f$ with $|\mathcal{A}_f|=\rho |\tilde{\mathcal{D}}_f|$.
    \EndFor

    \State Clinical reviewers independently assess sampled items for update-target alignment, vignette plausibility, answer correctness, rationale consistency, and clinically meaningful distractors or alternatives.
    \State Aggregate reviewer labels and adjudicate disagreements.
    \State Summarize recurrent failure modes and update the rendering prompts and constraints.
\Until{review pass rates stabilize across formats}

\State Conduct a final expert audit on the revised candidate corpora.
\State Revise correctable failures and remove unrecoverable failures.
\State Return the verified corpora $\{\mathcal{D}_f\}$ used for supervised fine-tuning.
\end{algorithmic}
\end{algorithm}

\section{Candidate-Level View of Format Structure}
\label{sec:appendix-format-formalization}

This appendix provides an analytical view of how different supervision formats organize contrastive information. All models in our experiments are trained with the same token-level autoregressive SFT objective. The candidate-level formulation below is therefore not an additional training loss; it is a diagnostic abstraction for comparing which clinically relevant alternatives are co-presented within each training instance.

\paragraph{Autoregressive scoring.}
For an input $x$ and an answer string $y$, the autoregressive model assigns the length-normalized sequence score
\begin{equation}
\ell_\theta(x,y)
=
\frac{1}{|y|}
\sum_{t=1}^{|y|}
\log p_\theta(y_t \mid x,y_{<t}) ,
\end{equation}
Given an analytical candidate set $\mathcal{Y}$, these scores induce a candidate-level distribution
\begin{equation}
p_\theta(y\mid x,\mathcal{Y})
=
\frac{\exp \ell_\theta(x,y)}
{\sum_{y'\in\mathcal{Y}}\exp \ell_\theta(x,y')} ,
\end{equation}
This distribution is used only to formalize how answer strings would compete under the same autoregressive model when they are considered together.

\paragraph{Co-presented alternatives.}
For update item $i$, let $a_{\mathrm{new},i}$ denote the current guideline-supported recommendation. Let $\mathcal{N}_i^{(f)}$ denote the non-current or clinically plausible alternatives exposed by format $f$ through the vignette, rationale, answer options, blank completion context, or explanation. The co-presented contrast set for item $i$ under format $f$ is
\begin{equation}
\mathcal{Z}_i^{(f)}
=
\{a_{\mathrm{new},i}\}\cup \mathcal{N}_i^{(f)} ,
\end{equation}

For \textsc{SAQ}, \textsc{FITB}, and \textsc{MSQ}, this contrast is local to a single clinical vignette:
\begin{equation}
\mathcal{Y}_i^{(f)}=\mathcal{Z}_i^{(f)},
\qquad
f\in\{\textsc{SAQ},\textsc{FITB},\textsc{MSQ}\},
\end{equation}
The formats differ in how the local alternatives are expressed: \textsc{SAQ} exposes them through the rationale, \textsc{FITB} through the masked entity and explanation, and \textsc{MSQ} through item-specific answer options.

\textsc{EMQ} changes the scope of co-presentation. For a block $G$ of related update items, multiple vignettes are presented together with a shared answer space. The analytical contrast set for each vignette in the block is therefore
\begin{equation}
\mathcal{Y}_{i,G}^{(\textsc{EMQ})}
=
\bigcup_{j\in G}\mathcal{Z}_j^{(\textsc{EMQ})},
\qquad i\in G .
\end{equation}
Thus each vignette is interpreted not only against its own local alternatives, but also against clinically related alternatives from adjacent update items in the same block.

\paragraph{Representation-space surrogate.}
Let $h_\theta(x)\in\mathbb{R}^d$ be the hidden representation of input $x$, and let $e_\theta(y)\in\mathbb{R}^d$ be a representation of answer string $y$. We define a compatibility score
\begin{equation}
s_\theta(x,y)=h_\theta(x)^\top e_\theta(y),
\end{equation}
For a candidate set $\mathcal{Y}$ containing the correct current recommendation $a_{\mathrm{new},i}$, the corresponding surrogate contrastive loss is
\begin{equation}
\widetilde{\mathcal{L}}(i;\mathcal{Y})
=
-\log
\frac{\exp s_\theta(x_i,a_{\mathrm{new},i})}
{\sum_{y\in\mathcal{Y}}\exp s_\theta(x_i,y)} ,
\end{equation}
For an \textsc{EMQ} block $G$, this gives
\begin{equation}
\widetilde{\mathcal{L}}_{\textsc{EMQ}}(G)
=
\frac{1}{|G|}
\sum_{i\in G}
\widetilde{\mathcal{L}}
\left(i;\mathcal{Y}_{i,G}^{(\textsc{EMQ})}\right),
\end{equation}

\paragraph{Gradient interpretation.}
Let $p_\theta(y\mid x_i,\mathcal{Y})$ denote the softmax distribution induced by the surrogate compatibility scores. The gradient of $\widetilde{\mathcal{L}}(i;\mathcal{Y})$ with respect to the query representation is
\begin{equation}
\begin{aligned}
\nabla_{h_\theta(x_i)}
\widetilde{\mathcal{L}}(i;\mathcal{Y})
&=
\sum_{y\in\mathcal{Y}}
p_\theta(y\mid x_i,\mathcal{Y})e_\theta(y) \\
&\quad
-
e_\theta(a_{\mathrm{new},i}) ,
\end{aligned}
\end{equation}
Equivalently,
\begin{equation}
\begin{aligned}
\nabla_{h_\theta(x_i)}
\widetilde{\mathcal{L}}(i;\mathcal{Y})
&=
\sum_{y\in\mathcal{Y}\setminus\{a_{\mathrm{new},i}\}}
p_\theta(y\mid x_i,\mathcal{Y}) \\
&\quad \cdot
\left(e_\theta(y)-e_\theta(a_{\mathrm{new},i})\right),
\end{aligned}
\end{equation}
This expression shows that the representation update is shaped by the alternatives co-presented with the current recommendation. Local formats expose alternatives tied to one vignette, whereas \textsc{EMQ} exposes alternatives across related vignettes in the same block. The resulting supervision signal is therefore broader and more relational, because clinically similar recommendations are contrasted within a shared instance.
\paragraph{Connection to diagnostics.}
This view motivates the diagnostics in Section~\ref{sec:analysis}. If \textsc{EMQ}'s advantage comes from networked clinical contrast rather than merely from longer text, then \textsc{EMQ} should expose more clinically relevant entities, place related entities closer together, and improve transfer under a fixed generation budget. If the shared structure guides updates toward task-relevant distinctions, it should also preserve discriminative representations with less unnecessary movement from the base model. The signal-density and representation analyses test these observable implications.

\section{Vignette and Token-Budget Controls}
\label{sec:appendix-vignette-token-controls}
This appendix examines whether the advantage of \textsc{EMQ} can be attributed mainly to larger vignette exposure or prompt-token efficiency. All supervision formats are rendered from the same pool of 2{,}530 unique clinical vignettes. Thus, the comparison uses the same vignette pool across formats; what differs is how the vignettes and their associated update contrasts are organized within training instances.

\paragraph{Vignette and token accounting.}
Table~\ref{tab:vignette-token-accounting} summarizes the training-token and vignette accounting for each format. We report the average number of training tokens per item, the average number of vignettes per item, the number of unique vignettes covered by the format, and the average prompt-token cost per vignette. Let $\bar{v}_f$ be the average number of vignettes per item and $\bar{t}_f$ be the average number of prompt tokens per item for format $f$. We define vignette density as
\begin{equation}
\label{eq:vignette-density}
\mathrm{VD}(f)
=
1000\,
\frac{\bar{v}_f}{\bar{t}_f},
\end{equation}
where higher values indicate that more clinical vignettes are exposed per 1{,}000 prompt tokens.

\begin{table*}[t]
\centering
\small
\setlength{\tabcolsep}{4pt}
\renewcommand{\arraystretch}{1.10}
\begin{tabular}{lccccc}
\toprule
\textbf{Format} &
\textbf{Avg. prompt tok./item} &
\textbf{Avg. vign./item} &
\textbf{Unique vign.} &
\textbf{Prompt tok./vign.} &
\textbf{Vign./1k prompt tok.} \\
\midrule
\textsc{SAQ}  & 261.5 & 1.00 & 2{,}530 & 261.5 & 3.82 \\
\textsc{MSQ}  & 260.2 & 1.00 & 2{,}530 & 260.2 & 3.84 \\
\textsc{FITB} & 254.3 & 1.00 & 2{,}530 & 254.3 & 3.93 \\
\textsc{EMQ}  & 258.7 & 2.72 & 2{,}530 & 95.1  & 10.51 \\
\bottomrule
\end{tabular}
\caption{Token and vignette accounting across supervision formats. All formats are rendered from the same pool of 2{,}530 unique vignettes. Vignette density measures how efficiently a format exposes clinical vignettes under a fixed prompt-token budget.}
\label{tab:vignette-token-accounting}
\end{table*}

\begin{table*}[t]
\centering
\small
\setlength{\tabcolsep}{4pt}
\renewcommand{\arraystretch}{1.10}
\begin{tabular}{lcccccc}
\toprule
\textbf{Format} &
\textbf{Train} &
\textbf{Exposed} &
\textbf{Train tok.} &
\textbf{Tok.} &
\textbf{SEER} &
\textbf{Rat.} \\
&
\textbf{items} &
\textbf{vign.} &
\textbf{(M)} &
\textbf{mult.} &
\textbf{Acc.} &
\textbf{Acc.} \\
\midrule
Expanded \textsc{SAQ}  & 6{,}882 & 6{,}882 & 4.56 & 2.74$\times$ & 63.4 & 53.4 \\
Expanded \textsc{MSQ}  & 6{,}882 & 6{,}882 & 4.50 & 2.71$\times$ & 62.9 & 56.2 \\
Expanded \textsc{FITB} & 6{,}882 & 6{,}882 & 4.43 & 2.66$\times$ & 62.2 & 51.6 \\
Full \textsc{EMQ}      & 2{,}530 & 6{,}882 & 1.66 & 1.00$\times$ & \textbf{64.8} & \textbf{59.6} \\
\bottomrule
\end{tabular}
\caption{Exposure-matched scaling control. The single-vignette formats are expanded to 6{,}882 training instances, matching the approximate number of vignette-level exposures in the full \textsc{EMQ} corpus. Token multiplier is computed relative to full \textsc{EMQ}.}
\label{tab:exposure-matched-scaling-control}
\end{table*}

The accounting shows that \textsc{EMQ} does not introduce additional unique clinical information. Instead, it packages multiple related vignettes into shared-answer training instances, allowing the model to integrate more vignette-level contrasts per unit of prompt budget.

\paragraph{Exposure-matched scaling control.}
We further test whether the \textsc{EMQ} advantage can be matched by simply increasing the amount of single-vignette supervision. Since the full \textsc{EMQ} corpus contains 2{,}530 training instances with an average of 2.72 vignettes per instance, it exposes approximately 6{,}882 vignette-level decisions in total. We therefore construct expanded \textsc{SAQ}, \textsc{MSQ}, and \textsc{FITB} corpora with 6{,}882 training instances each, while preserving their original single-vignette rendering rules. These expanded controls match \textsc{EMQ}'s vignette-level exposure but require substantially more supervised training tokens because each vignette is serialized as a separate training item.

Table~\ref{tab:exposure-matched-scaling-control} reports the resulting token cost and SEER-Bench performance. Total training tokens count both the prompt and supervised target tokens. Even after expanding the single-vignette formats to 6{,}882 instances, none matches the full \textsc{EMQ} result. The expanded controls require 2.66--2.74 times more training tokens than \textsc{EMQ}, yet remain lower on both SEER-Bench answer accuracy and rationale accuracy.

Together, these controls show that \textsc{EMQ}'s advantage is not explained by larger unique clinical content or by greater vignette-level exposure alone. Matching \textsc{EMQ}'s vignette exposure with single-vignette formats requires substantially more training tokens and still yields weaker SEER-Bench transfer. This supports the interpretation that \textsc{EMQ} is more token-efficient because it organizes related clinical vignettes and answer contrasts within shared training instances, rather than presenting each vignette as an isolated decision.

\section{SEER-Bench Construction}
\label{sec:appendix-seer}

SEER-Bench is designed for temporal evaluation of medical knowledge: we curate cases from the latest versioned SEER Research Data release ~\citep{seer2025}. This appendix documents the cohort filtering, cancer-site distribution, and quality-control details.

\subsection{Construction Pipeline}
\label{sec:seer-construction}

The construction proceeds through three stages:

\textbf{Stratified raw sampling and initial filtering.} We first construct a raw candidate pool by randomly sampling SEER patient records across the 16 targeted cancer-site groups. To ensure data viability prior to generation, we apply strict initial exclusion criteria: (1) cases with any missing or unknown NCCN staging variable (T, N, M, or Overall Stage); (2) histopathologic types that cannot be unambiguously mapped to the NCCN staging scheme\citep{nccn2025}; and (3) cases that are in situ or non-invasive. This filtering yields a pre-review pool of 3,200 qualified raw candidates (200 cases per cancer-site group), representing the candidate pool before text synthesis and expert validation.

\textbf{Vignette synthesis and item generation.} Both steps are performed jointly in a single Gemini 3.1 Pro call (full prompt in Figure\ref{fig:seer-prompt} ). The model receives coded SEER patient records (including cancer size code tables and regional node coding rules), converts them into coherent clinical narratives, randomly omits one staging variable (T, N, M, or Overall Stage) as the inference target, and provides an answer with rationale grounded in NCCN criteria. The synthesis is carefully constrained to prevent label leakage while ensuring the narrative contains sufficient clinical evidence for staging derivation.

\textbf{ Physician review and final inclusion.} Physician reviewers then rigorously screen the 3,200 generated items. Crucially, the gold labels are strictly derived from the structured fields of the source SEER records. An item is permanently discarded if: (1) any conflict arises between the model-generated text and these structured gold standard fields; or (2) the underlying SEER database records lack sufficient clinical details to robustly support a definitive TNM staging diagnosis. Following this expert filtering process, the final SEER-Bench dataset comprises 1,992 highly reliable cases across the 16 cancer-site groups.

\subsection{Statistics and Quality Control}
\label{sec:seer-stats}

\begin{figure}[t]
\centering
\includegraphics[width=\columnwidth]{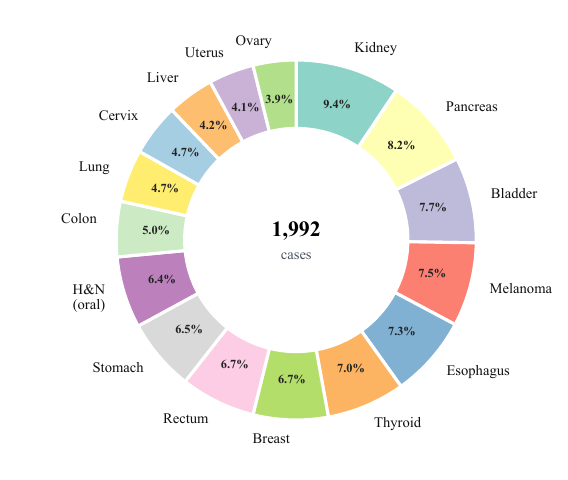}
\caption{Cancer-site distribution in SEER-Bench. Donut slices report the percentage of cases in the 1,992-case curated subset. The H\&N (oral) group aggregates oral-cavity sub-sites; the Colon and Rectum categories are reported separately.}
\label{fig:seer-site-distribution}
\end{figure}

Figure~\ref{fig:seer-site-distribution} summarizes the cancer-site composition of the curated subset. The distribution is intentionally broad rather than dominated by a single malignancy, which helps test whether updated models transfer staging knowledge across heterogeneous oncology settings.

This design keeps the benchmark temporally explicit: all items come from the same diagnosis-year cohort, each case masks exactly one staging target, and the final set is a physician-reviewed subset of the 3{,}200 stratified random candidates. As a result, the benchmark separates final-label prediction from rationale-level reasoning without mixing multiple temporal anchors.

Quality control (QC) is strictly enforced to mitigate specific dataset failure modes: (i) \emph{source traceability}, maintaining unbroken provenance to versioned SEER data; (ii) \emph{answer-space validation}, ensuring all target labels are grammatically and medically valid NCCN categories; (iii) \emph{omission consistency}, programmatically verifying that the masked variable remains unambiguously inferable from the unmasked text; and (iv) \emph{full double-blinded physician review}, in which physician reviewers are \textbf{randomly assigned to independent review panels} and conduct a stratified 100\% review of all candidate items by cancer-site group and masked variable. All reviews are performed double-blind to eliminate annotation bias, verifying consistency among the generated vignette, gold answer, and rationale. Only items that pass this rigorous expert review are included in the final 1{,}992-case SEER-Bench set; failed items are permanently excluded. 
Across physician verification labels, raw inter-rater agreement was 96.9\%, with Cohen's $\kappa=0.87$.

\subsection{Evaluation Protocol}
\label{sec:seer-eval}

SEER-Bench reports two metrics. \emph{Answer accuracy} measures whether the extracted final staging/classification label matches the ground-truth masked target after case-insensitive normalization and common semantic variants (e.g., ``Stage IA'' and ``Stage 1A''). \emph{Rationale accuracy} is stricter: a response is correct only when the final answer is correct and the explanation demonstrates the same key clinical reasoning as the reference rationale, including the relevant T/N/M logic, staging criteria, and interpretation of the clinical findings. The GPT-5.4 judge achieved 98.2\% raw agreement with human annotations, corresponding to Cohen's $\kappa=0.88$.

Let $m_i$ be the model response, $a_i=\operatorname{ans}(m_i)$ the extracted answer, $g_i$ the ground-truth answer, and $r_i$ the rationale-correctness judgment produced by GPT-5.4 using the grading prompt in Figure~\ref{fig:seer-rationale-prompt}. Operationally, \(\mathrm{Acc}\) averages \(\mathbf{1}[a_i=g_i]\) over all items, and \(\mathrm{RatAcc}\) averages \(\mathbf{1}[a_i=g_i \wedge r_i=1]\). If the final answer is incorrect, rationale correctness is forced to false.

\begin{figure*}[t]
\centering
\begin{tcolorbox}[
    colback=gray!3!white,
    colframe=gray!70!black,
    title=\textbf{SEER-Bench Vignette Generation Prompt},
    fonttitle=\bfseries\large,
    fontupper=\small,
    rounded corners,
    boxrule=0.8pt,
    arc=3mm,
    boxsep=3mm,
    toptitle=2mm,
    bottomtitle=2mm,
    before upper={\renewcommand\baselinestretch{1.1}\parskip=0.3em}
]
\textbf{Role:} Oncology Board Exam Developer and Clinical Educator. Synthesize high-quality clinical vignettes based on the SEER database to train medical students and residents on TNM staging and cancer prognosis.

\textbf{Input Data Format:} Patient records with: Gender, Age, Cancer Location, Pathological Diagnosis, Tumor Size (mm), Cancer Extension, Lymph Node Examined/Positive Status, Bone/Brain/Liver/Lung Metastatic Status, T/N/M/overall Stage.

\textbf{Input Data Coding:}
\begin{itemize}[leftmargin=1.2em, topsep=0pt, itemsep=0pt]
    \item \emph{Cancer Size}: coded 0 (no mass), 001--988 (exact mm), 989 ($\geq$989\,mm), 990 (microscopic focus), 991--995 (range descriptions), 999 (unknown).
    \item \emph{Regional nodes examined}: coded as the exact number of nodes examined; exceptions include aspiration (coded 95), combinations of positive aspirated/biopsied/sampled/dissected nodes (coded 98), and unknown/not applicable (coded 99).
    \item \emph{Regional nodes positive}: coded as the exact number of positive nodes; exceptions include positive aspiration (coded 95), combinations (coded 97), and cases where no nodes were examined (coded 98).  
\end{itemize}

\textbf{Task Instructions:}
\begin{enumerate}[leftmargin=1.2em, topsep=0pt, itemsep=0pt]
    \item \emph{Scenario Generation}: Convert the provided data into a professional clinical vignette. Do not just list the facts; describe the patient's presentation.
    \item \emph{The ``Omission'' Logic}: For every record, randomly select ONE of the following variables to omit from the vignette and turn into the ``Question'': the T-category, the N-category, the M-category, or the Overall Stage. Ask a clear lead-in question about the omitted variable.
    \item \emph{Answer \& Rationale}: Provide the correct answer and a brief explanation using the latest NCCN guidelines.
\end{enumerate}

\textbf{Output:} JSON with fields: \texttt{Question}, \texttt{Response}, \texttt{Rationale}.
\end{tcolorbox}
\caption{Prompt template for SEER-Bench vignette generation and item construction. The model receives coded SEER patient records, synthesizes clinical narratives, and constructs masked staging inference items.}
\label{fig:seer-prompt}
\end{figure*}

The vignette-generation prompt is where structured SEER codes are converted into natural-language clinical cases. Its constraints reduce label leakage while still requiring enough clinical evidence for the omitted T, N, M, or overall-stage target to be inferred.

\begin{figure*}[t]
\centering
\begin{tcolorbox}[
    colback=gray!3!white,
    colframe=gray!70!black,
    title=\textbf{SEER-Bench Rationale Evaluation Prompt},
    fonttitle=\bfseries\large,
    fontupper=\small,
    rounded corners,
    boxrule=0.8pt,
    arc=3mm,
    boxsep=3mm,
    toptitle=2mm,
    bottomtitle=2mm,
    before upper={\renewcommand\baselinestretch{1.1}\parskip=0.3em}
]
\textbf{Role:} Expert medical AI evaluation assistant. Evaluate whether a model response demonstrates both a correct final answer and correct reasoning/rationale.

\textbf{Inputs:}
\begin{itemize}[leftmargin=1.2em, topsep=0pt, itemsep=0pt]
    \item Ground Truth Answer: \texttt{\{gt\_answer\}}
    \item Ground Truth Rationale: \texttt{\{gt\_rationale\}}
    \item Model Response: \texttt{\{model\_response\}}
\end{itemize}

\textbf{Evaluation Rules:}
\begin{enumerate}[leftmargin=1.2em, topsep=0pt, itemsep=0pt]
    \item Extract the model's final answer or conclusion from its response.
    \item Check whether the final answer matches the ground-truth answer, case-insensitively and allowing semantically equivalent variants such as ``Stage IA'', ``Stage 1A'', and ``stage ia''.
    \item If and only if the answer is correct, evaluate whether the reasoning aligns with the ground-truth rationale. The wording need not be identical, but the response must demonstrate the same key clinical reasoning points, such as T/N/M classification logic, staging-criteria application, and interpretation of clinical findings.
    \item Mark the rationale correct if it identifies the same critical factors and logical steps. Minor omissions of non-essential details are acceptable. If the model reaches the right answer through clearly wrong reasoning, mark the rationale incorrect.
\end{enumerate}

\textbf{Output Requirement:} Output strictly in JSON with no other text:

\begin{quote}
\ttfamily\scriptsize
\{\\
\hspace*{1em}"extracted\_answer": "The final answer extracted from model response",\\
\hspace*{1em}"answer\_correct": true or false,\\
\hspace*{1em}"rationale\_analysis": "Brief explanation of whether the model's reasoning matches the ground truth rationale",\\
\hspace*{1em}"rationale\_correct": true or false\\
\}
\end{quote}

\textbf{Constraint:} If \texttt{answer\_correct} is false, \texttt{rationale\_correct} must also be false.
\end{tcolorbox}
\caption{Prompt template for SEER-Bench rationale-accuracy grading. The judge first verifies the final staging answer, then evaluates rationale correctness only for answer-correct responses.}
\label{fig:seer-rationale-prompt}
\end{figure*}

The rationale grader makes rationale accuracy stricter than answer accuracy: once the final staging label is wrong, the rationale is automatically treated as incorrect. This prevents fluent but clinically misaligned explanations from being counted as successful reasoning.
\section{Experimental Setup}
\label{sec:appendix-exp-setup}

\paragraph{Training data.} The main English experiments construct 2{,}530 SFT instances from the 2026 NCCN oncology guidelines for each of the four supervision formats (\textsc{EMQ}, \textsc{MSQ}, \textsc{FITB}, and \textsc{SAQ}). Each model variant is updated exclusively on the 2{,}530-instance corpus corresponding to its assigned format. A separate supplementary Chinese transfer check uses 168 SFT instances constructed from recent updated public vaccine guidance\citep{prevention2026expert,group2024practice}. Within each setting, the four supervision formats share the same update content and differ only in item structure.

\paragraph{Base model and adaptation.} The base model is Qwen3-4B (qwen3-4B-2507). We evaluate SFT-LoRA and trained on the update content rendered into \textsc{EMQ}, \textsc{MSQ}, \textsc{FITB}, and \textsc{SAQ}. Within the adaptation route, we use the same training budget and LoRA configuration across formats.

\paragraph{Comparison methods.} We compare the unmodified Qwen3-4B base model with retrieval-based updating (Base RAG and CARE), knowledge-editing baselines (RECIPE and AlphaEdit), format-controlled LoRA updates, and larger reference LLMs. The Base RAG inference prompt is provided in Figure~\ref{fig:prompt-rag}; method-specific hyperparameters are reported in Tables~\ref{tab:lora-rag-hparams}--\ref{tab:care-hparams}.

\paragraph{Evaluation benchmarks.} MedGUIDE-English contains HCC and NSCLC update slices built from historical NCCN guideline in 2024. SEER-Bench evaluates temporally anchored oncology staging with answer accuracy and rationale accuracy, defined in Appendix~\ref{sec:seer-eval}. HealthBench Professional evaluates external professional medical-QA generalization. The auxiliary Chinese transfer evaluation is reported separately and uses the vaccine-related subset of LLM-EvalMed (39 items) and the vaccine-related subset of LiveMedBench (204 items).

\paragraph{Automatic scoring and LLM judges.} To keep comparisons controlled within each benchmark, we fix the same judge across all evaluated models and methods for a given benchmark. SEER-Bench rationale accuracy are graded with GPT-5.4 and MedGUIDE-English with Gemini 3.1 Pro; HealthBench Professional, LiveMedBench, and LLM-EvalMed are graded with GPT-5.4. For Chinese benchmarks whose original reports use different judges, LiveMedBench with GPT-OSS and LLM-EvalMed with GPT-4o.

Table~\ref{tab:model_details} lists the foundation and API models used for updating, cross-scale validation, reference evaluation, data generation, and automatic grading. Qwen3-4B is the main controlled backbone; Qwen3-1.7B, Qwen3-8B, and Qwen3-14B are used for the cross-scale SEER-Bench check; Llama-3.1-8B-Instruct is used as an additional backbone validation; and MediTron-70B and HuatuoGPT-O1-70B are included as medical reference systems. API-based models were accessed via their default inference endpoints with no custom system prompts for evaluation; temperature was set to 0 when the endpoint exposes this parameter, and default settings were used otherwise. Open-weight experiments were run locally in bf16. References point to technical reports, official model cards, or provider documentation.

\begin{table*}[!t]
\small
\centering
\caption{Full model specifications for the foundation and API models used in the experiments.}
\label{tab:model_details}
\resizebox{\textwidth}{!}{
\begin{tabular}{llllllc}
\toprule
\textbf{Model} & \textbf{Provider} & \textbf{Version / Checkpoint} & \textbf{Params} & \textbf{Access} & \textbf{Role} & \textbf{Ref.} \\
\midrule
Qwen3-1.7B & Alibaba & Qwen/Qwen3-1.7B & 1.7B & HF/local & Cross-scale backbone & \citep{qwen2025qwen317b} \\
Qwen3-4B & Alibaba & Qwen/Qwen3-4B-Instruct-2507 & 4.0B & HF/local & Main/update backbone & \citep{qwen2025qwen34b2507} \\
Qwen3-8B & Alibaba & Qwen/Qwen3-8B & 8.2B & HF/local & Cross-scale backbone & \citep{qwen2025qwen38b} \\
Qwen3-14B & Alibaba & Qwen/Qwen3-14B & 14.8B & HF/local & Cross-scale backbone & \citep{qwen2025qwen314b} \\
Qwen3-235B & Alibaba & Qwen/Qwen3-235B-A22B-Instruct-2507 & 235B & API & Reference system & \citep{qwen2025qwen3235b2507} \\
\midrule
Llama-3.1-8B-Instruct & Meta & meta-llama/Llama-3.1-8B-Instruct & 8B & HF/local & Backbone validation & \citep{grattafiori2024llama} \\
MediTron-70B & EPFL & epfl-llm/meditron-70b & 70B & HF/local & Medical reference & \citep{chen2023meditron} \\
HuatuoGPT-O1-70B & FreedomIntelligence & FreedomIntelligence/HuatuoGPT-o1-70B & 70B & HF/local & Medical reference & \citep{chen2025towards} \\
\midrule
Kimi-2.5 & Moonshot AI & kimi-k2.5 & -- & API & Reference system & \citep{moonshot2026kimik25} \\
GLM-5 & Zhipu AI & glm-5 & -- & API & Reference system & \citep{zeng2026glm} \\
DeepSeek-R1 & DeepSeek & deepseek-r1 & 671B & API & Reference system & \citep{guo2025deepseek} \\
Claude-Opus-4.6 & Anthropic & claude-opus-4-6 & -- & API & Reference system & \citep{anthropic2026claudeopus46} \\
Gemini-3.1-Pro & Google DeepMind & gemini-3.1-pro & -- & API & Reference/generation/judge & \citep{deepmind2026gemini31pro} \\
GPT-5.4 & OpenAI & gpt-5.4 & -- & API & Reference/judge & \citep{openai2026gpt54} \\
\bottomrule
\end{tabular}
}
\end{table*}

The table separates the roles of backbone model, medical reference model, general reference model, data-generation model, and automatic judge. Controlled rows share the same foundation checkpoint within each backbone, while external reference systems provide descriptive context rather than directly comparable training-controlled variants.

\subsection{Implementation Hyperparameters}
\label{sec:appendix-hyperparams}

Tables~\ref{tab:lora-rag-hparams}--\ref{tab:care-hparams} list the implementation hyperparameters used for the reported LoRA, retrieval, and knowledge-editing baselines. Within each adaptation route, the same settings are held fixed across supervision formats.

\begin{table}[t]
\centering
\scriptsize
\setlength{\tabcolsep}{4pt}
\renewcommand{\arraystretch}{1.08}
\begin{tabularx}{\columnwidth}{llX}
\toprule
\textbf{Method} & \textbf{Hyperparameter} & \textbf{Value} \\
\midrule
\multirow{5}{*}{SFT-LoRA}
& LoRA rank / alpha / dropout & 128 / 256 / 0.1 \\
& Training epochs & 3 \\
& Batch size & 16 \\
& Warmup ratio & 0.05 \\
& Learning rate & 5e-5 \\
\midrule
\multirow{2}{*}{Base RAG}
& Embedding model & MedEmbed-large-v0.1 \\
& Retrieved passages & Top 3 \\
\bottomrule
\end{tabularx}
\caption{Hyperparameters for LoRA-based adaptation and the Base RAG retrieval baseline.}
\label{tab:lora-rag-hparams}
\end{table}

The LoRA settings are fixed within each adaptation route, and the retrieval baseline uses the same encoder and top-$k$ setting across questions. Consequently, differences among \textsc{EMQ}, \textsc{MSQ}, \textsc{FITB}, and \textsc{SAQ} are intended to reflect supervision structure rather than extra tuning budget.

\begin{table}[t]
\centering
\scriptsize
\setlength{\tabcolsep}{3pt}
\renewcommand{\arraystretch}{1.08}
\begin{tabularx}{\columnwidth}{lX}
\toprule
\textbf{Hyperparameter} & \textbf{Value} \\
\midrule
\multicolumn{2}{l}{\textbf{AlphaEdit}} \\
\texttt{layers} & [4, 5, 6, 7, 8] \\
\texttt{v\_num\_grad\_steps} & 25 \\
\texttt{v\_lr} & 1e-1 \\
\texttt{v\_loss\_layer} & 35 \\
\texttt{v\_weight\_decay} & 0.5 \\
\texttt{clamp\_norm\_factor} & 0.75 \\
\texttt{kl\_factor} & 0.0625 \\
\texttt{mom2\_adjustment} & true \\
\texttt{mom2\_update\_weight} & 15000 \\
\texttt{mom2\_n\_samples} & 10{,}000 \\
\texttt{rewrite\_module\_tmp} & \texttt{model.layers.\{\}.mlp.down\_proj} \\
\texttt{lm\_head\_module} & \texttt{model.embed\_tokens} \\
\texttt{nullspace\_threshold} & 2e-2 \\
\texttt{L2} & 10 \\
\midrule
\multicolumn{2}{l}{\textbf{RECIPE}} \\
\texttt{edit\_model\_name} & \texttt{qwen3-4b} \\
\texttt{model\_hidden\_size} & 2560 \\
\texttt{knowledge\_rep\_dim} & 2560 \\
\texttt{know\_rep\_prot\_token\_n} & 10 \\
\texttt{prompt\_token\_n} & 3 \\
\texttt{krm\_lr} & 1e-5 \\
\texttt{pt\_lr} & 1e-5 \\
\texttt{batch\_size} & 2 \\
\texttt{grad\_accum\_steps} & 8 \\
\texttt{max\_epochs} & 1000 \\
\texttt{early\_stop\_patience} & 7 \\
\texttt{contra\_lambda} / \texttt{relia\_lambda} & 2 / 1 \\
\bottomrule
\end{tabularx}
\caption{Hyperparameters for the AlphaEdit and RECIPE knowledge-editing baselines on Qwen3-4B. For RECIPE, \texttt{pt\_lr} is listed for completeness but the implementation uses \texttt{krm\_lr}; batch size 2 with 8 accumulation steps gives an effective batch size of 16.}
\label{tab:editing-hparams}
\end{table}

These settings document the editing baselines as reproducibility controls. They also clarify that the comparison is between different update mechanisms, not between independently optimized hidden-size or batch-size choices.

\begin{table}[t]
\centering
\scriptsize
\setlength{\tabcolsep}{3pt}
\renewcommand{\arraystretch}{1.08}
\begin{tabularx}{\columnwidth}{lX}
\toprule
\textbf{Hyperparameter} & \textbf{Value} \\
\midrule
\multicolumn{2}{l}{\textbf{CARE pretraining}} \\
Learning rate & 2e-4 \\
Number of training epochs & 1 \\
Total batch size & 384 (8 $\times$ 8 $\times$ 6) \\
Max sequence length & 336 \\
Retrieval context length & 180 \\
ICAE memory size & 16 \\
Memory hidden size & 2560 \\
\(\alpha_{\mathrm{NLL}}\) & 1.0 \\
LR scheduler / warmup ratio & linear / 0.03 \\
Weight decay & 0.0 \\
Seed & 980406 \\
LoRA rank / alpha / dropout & 8 / 16 / 0.05 \\
LoRA targets & \texttt{q/k/v/o/gate/up/down\_proj} \\
\midrule
\multicolumn{2}{l}{\textbf{CARE finetuning}} \\
Learning rate & 3e-4 \\
Number of training epochs & 1 \\
Total batch size & 48 (8 $\times$ 6 $\times$ 1) \\
Max sequence length & 512 \\
\(\alpha_{\mathrm{NLL}}\) / \(\alpha_{\mathrm{KL}}\) & 1.0 / 2.0 \\
KL temperature & 1.0 \\
ICAE memory size & 16 \\
Context teacher / student & adaptive / gt \\
Selection criterion / context selection & \texttt{closed\_book\_correct} / \texttt{nn} \\
\bottomrule
\end{tabularx}
\caption{CARE hyperparameters for Qwen3-4B. Pretraining uses 8 Ascend 910B devices; finetuning uses the best configuration from \texttt{run\_finetune\_full.sh}.}
\label{tab:care-hparams}
\end{table}

The CARE configuration is reported separately because it includes both pretraining and finetuning phases. This distinction matters when comparing it with LoRA-only updates, which do not add an additional retrieval-compression pretraining stage.
\section{Inference Prompt}
\label{sec:appendix-inference-prompts}

Figure~\ref{fig:prompt-rag} shows the inference prompt used for the Base RAG baseline. The prompt supplies the retrieved background knowledge and the target question, without exposing the supervision-format labels used in training.

\begin{figure*}[!t]
\centering
\begin{tcolorbox}[
    colback=gray!3!white,
    colframe=gray!70!black,
    title=\textbf{Base RAG Inference Prompt},
    fonttitle=\bfseries\large,
    fontupper=\small,
    rounded corners,
    boxrule=0.8pt,
    arc=3mm,
    boxsep=3mm,
    toptitle=2mm,
    bottomtitle=2mm,
    before upper={\renewcommand\baselinestretch{1.1}\parskip=0.3em}
]
\textbf{Role:} Medical expert assistant.

\textbf{Instruction:} Please answer the following question based on the provided background knowledge.

\textbf{Question:} \texttt{\{question\}}

\textbf{Background Knowledge:} \texttt{\{background\}}

\textbf{Response Requirement:} Provide a comprehensive and accurate answer based on the above information.
\end{tcolorbox}
\caption{Prompt template for the Base RAG baseline.}
\label{fig:prompt-rag}
\end{figure*}

The inference prompt intentionally exposes retrieved background knowledge but not the supervision-format labels used during training. This makes Base RAG a test-time knowledge-access baseline rather than another format-controlled update method.
\section{Auxiliary Chinese Evaluation}
\label{sec:appendix-chinese-eval}

Table~\ref{tab:chinese-prelim} reports an auxiliary Chinese transfer check on LLM-EvalMed and LiveMedBench, including both full-benchmark averages and vaccine-related subsets.

\begin{table*}[t]
\centering
\scriptsize
\setlength{\tabcolsep}{5pt}
\resizebox{\textwidth}{!}{%
\begin{tabular}{lcccc}
\toprule
\textbf{Model / method} &
\textbf{LLM-EvalMed-All}~$\uparrow$ &
\textbf{LLM-EvalMed-Sub.}~$\uparrow$ &
\textbf{LiveMedBench-Sub.}~$\uparrow$ &
\textbf{LiveMedBench-All}~$\uparrow$ \\
\midrule
Qwen3-4B & 3.479 & 3.558 & 0.4489 & 0.4023 \\
\midrule
SFT-LoRA + \textsc{EMQ} & \textbf{3.559} & \textbf{3.967} & \textbf{0.5111} & \textbf{0.4213} \\
SFT-LoRA + \textsc{MSQ} & 3.490 & 3.806 & 0.4860 & 0.3903 \\
SFT-LoRA + \textsc{FITB} & 3.459 & 3.664 & 0.4405 & 0.3873 \\
SFT-LoRA + \textsc{SAQ} & 3.519 & 3.750 & 0.4991 & 0.4133 \\
\midrule
Qwen3-235B & 3.814 & 4.114 & 0.5507 & 0.4616 \\
Kimi-2.5 & 3.841 & 3.897 & \textbf{0.5644} & 0.5518 \\
GLM-5 & 3.791 & 3.930 & 0.5208 & 0.5419 \\
DeepSeek-R1 & 3.933 & 3.943 & 0.5077 & 0.4697 \\
Claude-Opus-4.6 & 3.927 & 3.956 & 0.5634 & 0.5531 \\
Gemini-3.1-Pro & \textbf{3.975} & \textbf{4.297} & 0.5032 & \textbf{0.5596} \\
GPT-5.4 & 3.941 & 3.973 & 0.5203 & 0.4697 \\
\bottomrule
\end{tabular}
}
\caption{Auxiliary Chinese transfer evaluation results. This supplementary setting uses vaccine-guidance updates and is separate from the main NCCN English experiments. LLM-EvalMed-All reports the full 667-item average, and LLM-EvalMed-Sub. reports the 39-item vaccine-related subset. LiveMedBench-Sub. reports the vaccine-related subset, and LiveMedBench-All reports the full benchmark average. Both Chinese benchmarks are re-scored with GPT-5.4; original benchmark reports use different judges, so scores are not directly interchangeable.}
\label{tab:chinese-prelim}
\end{table*}

This auxiliary transfer check should be interpreted separately from the main oncology update evaluation. It uses vaccine-guidance updates and Chinese benchmarks, so it tests whether format-controlled updating carries cross-lingual signal rather than directly replicating the NCCN English setting. The large reference models provide external context rather than format-controlled variants of the same update procedure.
\section{Case Study: Knowledge Updating Challenges}
\label{sec:appendix-cases}

We present two representative cases from MedGUIDE that illustrate the dual challenges of medical knowledge updating: \emph{stability maintenance} and \emph{update learning}. For each case, we show the clinical vignette, the correct answers from the outdated and latest versions, and the key guideline change that distinguishes the two versions. Figures~\ref{fig:case-hcc} and~\ref{fig:case-nsclc} present the detailed case analyses. We then add a SEER-Bench renal cell carcinoma (RCC) staging case to illustrate the intended output-length and correctness pattern across the four supervision formats.

\begin{figure*}[t]
    \centering
    \begin{tcolorbox}[
        colback=gray!3!white,
        colframe=gray!70!black,
        title=\textbf{Case Study: Stability Maintenance (hcc-1)},
        fonttitle=\bfseries\large,
        fontupper=\footnotesize,
        rounded corners,
        boxrule=0.8pt,
        arc=3mm,
        boxsep=2mm,
        toptitle=2mm,
        bottomtitle=2mm,
        before upper={\renewcommand\baselinestretch{1.1}\parskip=0.3em}
    ]
        \textbf{\textcolor{gray!40!black}{1. Source \& Domain}} \hfill \textit{NCCN Hepatocellular Carcinoma (HCC) Screening}

        \vspace{0.2em}
        {\color{black!40}\hrule height 0.6pt}
        \vspace{0.3em}

        \textbf{\textcolor{gray!40!black}{2. Clinical Vignette}}

        \textit{A 43-year-old male with a history of hepatitis B and liver cirrhosis presented for follow-up after being treated for HCC with ablation therapy two years ago. His last ultrasound (US), performed six months ago, showed no signs of cancer recurrence, and his alpha-fetoprotein (AFP) level was normal. He has been feeling well, with no new symptoms, but remains at high risk for HCC recurrence due to his underlying liver disease. His liver function is stable, and he has maintained regular surveillance. Today, an ultrasound and AFP test are repeated as part of his ongoing monitoring protocol. The ultrasound findings are negative for any lesions, and his AFP levels remain within the normal range.}

        \vspace{0.2em}
        {\color{black!40}\hrule height 0.6pt}
        \vspace{0.3em}

        \textbf{\textcolor{gray!40!black}{3. Question}}

        \textit{Given these recent results and the necessity for continued vigilance in managing his condition, what should be the next step in his surveillance plan?}

        \vspace{0.2em}
        {\color{black!40}\hrule height 0.6pt}
        \vspace{0.3em}

        \textbf{\textcolor{gray!40!black}{4. Choices}}
        \begin{itemize}[leftmargin=1.5em, topsep=2pt, itemsep=0pt]
            \item[\textbf{A:}] Repeat US + AFP in 3--6 mo
            \item[\textbf{B:}] Additional workup (HCC-2)
            \item[\textbf{C:}] Repeat US + AFP in 6 mo
        \end{itemize}

        \vspace{0.2em}
        {\color{black!40}\hrule height 0.6pt}
        \vspace{0.3em}

        \textbf{\textcolor{gray!40!black}{5. Answers From Outdated and Latest Versions}}

        \begin{tcolorbox}[colback=teal!8, colframe=teal!50!black, boxrule=0.4pt, arc=1.5mm, left=4pt, right=4pt, top=3pt, bottom=3pt]
        \textbf{Answer from outdated version:} \colorbox{teal!20}{\textbf{C}} \quad \textit{(US negative $\rightarrow$ Repeat US + AFP in 6 mo)}

        \textbf{Answer from latest version:} \colorbox{teal!20}{\textbf{C}} \quad \textit{(Pathway preserved; \textbf{answer unchanged})}
        \end{tcolorbox}

        \vspace{0.2em}
        {\color{black!40}\hrule height 0.6pt}
        \vspace{0.3em}

        \textbf{\textcolor{gray!40!black}{6. Guideline Change ($\Delta$)}}

        The latest version updates the \emph{high-risk population definition}, but the downstream screening pathway for US-negative, AFP-normal patients remains \colorbox{yellow!30}{\emph{identical}}.

        \vspace{0.2em}
        {\color{black!40}\hrule height 0.6pt}
        \vspace{0.3em}

        \textbf{\textcolor{gray!40!black}{7. Test Target:}} \tcbox[on line, colback=teal!15, colframe=teal!60!black, boxrule=0.4pt, arc=1mm, boxsep=1pt, left=3pt, right=3pt]{\textbf{\textcolor{teal!70!black}{Stability}}}

        The model must \colorbox{teal!15}{\textbf{retain}} the correct answer (\textbf{C}) despite contextual guideline changes. Over-updating to A or B indicates failure to distinguish changed vs.\ unchanged knowledge.
    \end{tcolorbox}
    \caption{Case hcc-1: The guideline revision changes population criteria but not the screening pathway. A correctly updated model should maintain answer C across both versions.}
    \label{fig:case-hcc}
\end{figure*}

This case isolates stability: the surrounding guideline context changes, but the downstream answer should not. A model that changes away from C is not simply outdated; it is over-applying the update signal to an unchanged clinical decision path.

\begin{figure*}[t]
    \centering
    \begin{tcolorbox}[
        colback=gray!3!white,
        colframe=gray!70!black,
        title=\textbf{Case Study: Update Learning (nsclc-her2)},
        fonttitle=\bfseries\large,
        fontupper=\footnotesize,
        rounded corners,
        boxrule=0.8pt,
        arc=3mm,
        boxsep=2mm,
        toptitle=2mm,
        bottomtitle=2mm,
        before upper={\renewcommand\baselinestretch{1.1}\parskip=0.3em}
    ]
        \textbf{\textcolor{gray!40!black}{1. Source \& Domain}} \hfill \textit{NCCN NSCLC ERBB2 (HER2) Mutation First-Line Therapy}

        \vspace{0.2em}
        {\color{black!40}\hrule height 0.6pt}
        \vspace{0.3em}

        \textbf{\textcolor{gray!40!black}{2. Clinical Vignette}}

        \textit{A 61-year-old woman, a never-smoker, presents with progressive dyspnea, fatigue, and a 10-pound weight loss over the past two months. Imaging reveals a 4.5\,cm right lower lobe lung mass, a malignant pleural effusion, and multiple hepatic lesions. Biopsy confirms metastatic lung adenocarcinoma. Broad-panel molecular profiling of the cancer tissue is positive for an ERBB2 (HER2) mutation, with no other actionable driver alterations identified. She has not received prior systemic therapy for her cancer and maintains an ECOG performance status of 1.}

        \vspace{0.2em}
        {\color{black!40}\hrule height 0.6pt}
        \vspace{0.3em}

        \textbf{\textcolor{gray!40!black}{3. Question}}

        \textit{Given her treatment-naive status, what are the most appropriate first-line therapy options according to the latest version?}

        \vspace{0.2em}
        {\color{black!40}\hrule height 0.6pt}
        \vspace{0.3em}

        \textbf{\textcolor{gray!40!black}{4. Choices}}
        \begin{itemize}[leftmargin=1.5em, topsep=2pt, itemsep=0pt]
            \item[\textbf{A:}] Systemic therapy
            \item[\textbf{B:}] Zongertinib or systemic therapy
            \item[\textbf{C:}] Fam-trastuzumab deruxtecan-nxki or zongertinib
            \item[\textbf{D:}] Zongertinib or sevabertinib
            \item[\textbf{E:}] Ado-trastuzumab emtansine or systemic therapy
        \end{itemize}

        \vspace{0.2em}
        {\color{black!40}\hrule height 0.6pt}
        \vspace{0.3em}

        \textbf{\textcolor{gray!40!black}{5. Answers From Historical and Latest Versions}}

        \begin{tcolorbox}[colback=red!5, colframe=red!50!black, boxrule=0.4pt, arc=1.5mm, left=4pt, right=4pt, top=3pt, bottom=3pt]
        \textbf{Answer from historical version:} \colorbox{red!15}{\textbf{A}} \quad \textit{(ERBB2-mutated metastatic NSCLC routed to standard systemic therapy)}

        \textbf{Answer from latest version:} \colorbox{red!15}{\textbf{B}} \quad \textit{(Zongertinib is introduced as a preferred first-line option alongside systemic therapy)} \hfill \textbf{\textcolor{red!70!black}{Answer Changed!}}
        \end{tcolorbox}

        \vspace{0.2em}
        {\color{black!40}\hrule height 0.6pt}
        \vspace{0.3em}

        \textbf{\textcolor{gray!40!black}{6. Guideline Change ($\Delta$)}}

        The latest guideline iteration introduces a targeted therapeutic option into the first-line option for treatment-naive metastatic NSCLC harboring an ERBB2 (HER2) mutation. While historical versions restricted initial management to non-targeted systemic therapy, the updated protocol formally includes zongertinib or systemic therapy as appropriate first-line options.

        \vspace{0.2em}
        {\color{black!40}\hrule height 0.6pt}
        \vspace{0.3em}

        \textbf{\textcolor{gray!40!black}{7. Test Target:}} \tcbox[on line, colback=red!12, colframe=red!60!black, boxrule=0.4pt, arc=1mm, boxsep=1pt, left=3pt, right=3pt]{\textbf{\textcolor{red!70!black}{Plasticity}}}

        This case evaluates the model's capacity to successfully update its parametric knowledge and \textbf{shift its preferred option from A to B}. Because the historical answer (A) remains superficially clinically logical, this case serves as a highly challenging task to detect over-conservative model behavior.
    \end{tcolorbox}
    \caption{Case nsclc-her2: The guideline revision substantively changes the correct answer from A to B. A correctly updated model must adopt the latest recommendation.}
    \label{fig:case-nsclc}
\end{figure*}

This plasticity case tests whether the model can override the plausible outdated answer A and adopt B only when the guideline decision rule truly changes.

\newcommand{\rccok}[1]{\textbf{\textcolor{teal!65!black}{#1}}}
\newcommand{\rccbad}[1]{\textbf{\textcolor{red!70!black}{#1}}}
\newcommand{\rccgap}{\textcolor{black!50}{\ldots}}

\begin{figure*}[!t]
    \centering
    \begin{tcolorbox}[
        colback=gray!3!white,
        colframe=gray!70!black,
        title=\textbf{Case Study: Four-Format RCC Staging Output Excerpts},
        fonttitle=\bfseries\large,
        fontupper=\footnotesize,
        rounded corners,
        boxrule=0.8pt,
        arc=3mm,
        boxsep=2mm,
        toptitle=2mm,
        bottomtitle=2mm,
        before upper={\renewcommand\baselinestretch{1.08}\parskip=0.26em}
    ]
        \textbf{\textcolor{gray!40!black}{Shared Clinical Vignette.}}
        \textit{A 70-year-old non-Hispanic white male presents with a two-month history of left-sided flank pain and intermittent painless gross hematuria. Contrast-enhanced CT reveals a 9.0\,cm solid enhancing mass in the lower pole of the left kidney. He undergoes left radical nephrectomy and regional lymph-node dissection. Pathology shows ISUP grade 4 clear-cell renal cell carcinoma measuring 90\,mm in greatest dimension. The tumor invades the perinephric fat and has gross extension into the main left renal vein, but does not extend beyond Gerota's fascia. Four regional lymph nodes are examined and are negative. Staging chest CT and bone scan show no distant metastatic disease. What is the appropriate overall stage?}

        \textbf{\textcolor{gray!40!black}{Choices.}}
        A: Stage II; \quad B: Stage III; \quad C: Stage IV; \quad D: Stage I.

        \vspace{0.35em}
        {\color{black!40}\hrule height 0.6pt}
        \vspace{0.5em}

        \begin{tcolorbox}[colback=teal!4, colframe=teal!55!black, boxrule=0.45pt, arc=1.3mm, left=3pt, right=3pt, top=2pt, bottom=2pt]
        \textbf{\textsc{EMQ} excerpt (full output: 264 words):} \colorbox{teal!18}{\textbf{B correct}}\\
        To determine the appropriate overall stage, apply the applicable RCC TNM staging criteria. \rccgap{} The tumor \rccok{invades the perinephric fat and extends into the main left renal vein}, but does not extend beyond Gerota's fascia. \rccgap{} \rccok{T3a} includes renal-vein or perirenal-fat invasion without extension beyond Gerota's fascia. Four regional lymph nodes are negative, so \rccok{N0}; chest CT and bone scan show no distant disease, so \rccok{M0}. Therefore \rccok{T3aN0M0} maps to \rccok{Stage III}. \textbf{Final answer: B. Stage III.}
        \end{tcolorbox}

        \begin{tcolorbox}[colback=teal!4, colframe=teal!55!black, boxrule=0.45pt, arc=1.3mm, left=3pt, right=3pt, top=2pt, bottom=2pt]
        \textbf{\textsc{MSQ} excerpt (full output: 293 words):} \colorbox{teal!18}{\textbf{B correct}}\\
        The appropriate overall prognostic stage is Stage III. \rccgap{} The tumor measures 90\,mm (9.0\,cm), but the decisive feature is that it has \rccok{invaded the perinephric fat and the main left renal vein}. Although it remains within Gerota's fascia, this pattern is sufficient for a \rccok{T3a designation}. \rccok{N0} and \rccok{M0} are established by negative nodes and no distant metastasis, yielding \rccok{T3aN0M0} and \rccok{Stage III}. Option A is incorrect because Stage II corresponds to T2 disease confined to the kidney; despite the large size, venous and fat invasion upgrade the tumor. Options C and D are rejected by the absence of T4/M1 disease and by tumor size/extension.
        \end{tcolorbox}

        \begin{tcolorbox}[colback=teal!4, colframe=teal!55!black, boxrule=0.45pt, arc=1.3mm, left=3pt, right=3pt, top=2pt, bottom=2pt]
        \textbf{\textsc{SAQ} excerpt (full output: 432 words):} \colorbox{teal!18}{\textbf{B correct}}\\
        The tumor is 90\,mm, or 9.0\,cm, so size alone could suggest T2a. \rccok{But we cannot just look at size.} The tumor invades the perinephric fat and extends into the main left renal vein, so it is no longer limited to the kidney. \rccgap{} Because it \rccok{does not extend beyond Gerota's fascia}, renal-vein and perinephric-fat invasion make this \rccok{T3a}. Four nodes are negative, so \rccok{N0}; chest CT and bone scan are negative, so \rccok{M0}. Gathering the data gives \rccok{T3a, N0, M0}. Stage I is T1N0M0, Stage II is T2N0M0, and \rccok{Stage III includes T3N0M0}. Thus the answer is \rccok{Stage III}, \textbf{option B.}
        \end{tcolorbox}

        \begin{tcolorbox}[colback=red!4, colframe=red!55!black, boxrule=0.45pt, arc=1.3mm, left=3pt, right=3pt, top=2pt, bottom=2pt]
        \textbf{\textsc{FITB} excerpt (full output: 450 words):} \colorbox{red!15}{\textbf{A wrong}}\\
        The primary determinants are tumor size, nodal status, and distant metastasis. The key quantitative finding is that \rccbad{the tumor measures 9.0\,cm,} and the T category is determined \rccbad{primarily by the maximum diameter of the primary mass}. At 9.0\,cm, the tumor \rccbad{falls squarely within the T2a category}. Although perinephric-fat invasion and renal-vein involvement are reported, these findings should be regarded as \rccbad{secondary features that do not alter the size-based T staging}. The tumor is therefore assigned \rccbad{T2a}; together with negative regional lymph nodes and no distant metastases, this yields \rccbad{T2aN0M0}, corresponding to \rccbad{Stage II}. \textbf{Final answer: A. Stage II.}
        \end{tcolorbox}
    \end{tcolorbox}
    \caption{Representative RCC staging case rendered across four supervision formats. Each panel shows a highlighted excerpt, with parenthetical word counts reporting the full original output length. Green highlights mark the decisive T3aN0M0 reasoning preserved by \textsc{EMQ}, \textsc{MSQ}, and \textsc{SAQ}; red highlights mark the tumor-size shortcut that makes the \textsc{FITB} response select Stage II incorrectly.}
    \label{fig:case-rcc-formats}
\end{figure*}

\subsection{Analysis}

These cases jointly illustrate why supervision format matters for medical knowledge updating:

\indent Figure~\ref{fig:case-rcc-formats} holds the clinical vignette fixed and compares the outputs induced by the four supervision formats. The panels show highlighted excerpts, while the parenthetical word counts refer to the full original outputs. The example separates verbosity from correctness: \textsc{EMQ}, \textsc{MSQ}, and \textsc{SAQ} preserve the decisive renal-vein/perinephric-fat invasion cue and therefore recover T3aN0M0 and Stage III, whereas \textsc{FITB} produces the longest rationale but over-relies on the 9.0\,cm tumor size, consequently overlooking the vein invasion to misclassify the case as T2aN0M0, and selects Stage II.

\indent\textbf{Stability vs.\ Plasticity Dilemma.} Case~1 requires the model to \emph{maintain} its existing correct answer despite peripheral guideline revisions, while Case~2 requires the model to \emph{change} its answer when the recommendation genuinely shifts. Balancing these two demands is the core challenge of knowledge updating.

\indent\textbf{Failure Modes of Isolated Formats.} Under SAQ, MSQ, and FITB supervision, each training sample establishes only a single knowledge correspondence. This makes it difficult for models to learn the boundary between ``what has changed'' and ``what has not.'' Empirically, such models can under-update on Case~2 (persisting with the plausible outdated answer A), over-update on Case~1 (unnecessarily shifting away from the stable answer C), or produce long but incorrect slot-centered rationales as in the RCC \textsc{FITB} example.

\indent\textbf{Why EMQ Helps.} The \textsc{EMQ} format places multiple related clinical scenarios (including both changed and unchanged cases) within a single training instance, requiring the model to jointly assign each scenario to its correct answer from a shared candidate pool. This contrastive structure encourages simultaneous discrimination between stable and updated knowledge, making useful decision boundaries more likely to emerge.

\section{Clinical Decision Failure Modes Under Guideline Obsolescence}
\label{sec:appendix-clinical-safety}

Medical knowledge obsolescence can manifest as output-level failures with distinct downstream clinical consequences. These failure modes differ in how an LLM translates outdated knowledge into unsafe outputs. Incorrect staging can alter the model’s clinical conclusion and, in turn, redirect downstream treatment planning. Outdated treatment recommendations may preserve superseded interventions or treatment sequences, leading the model to recommend choices that are no longer supported by current guidelines. Overconfident rationales, by contrast, can make outdated recommendations harder to detect by presenting them with evidence that was valid under earlier clinical standards. Together, these failure modes motivate temporally anchored evaluation of both model answers and their rationales, as well as supervision that explicitly captures contrasts between historical and revised clinical conditions. Persistent underrepresentation of specific cancer types, languages, and guideline regions further poses a deployment risk, as limited exposure to relevant update events may reduce the model’s ability to recognize and correct knowledge that has become outdated.

\begin{table*}[t]
\centering
\scriptsize
\setlength{\tabcolsep}{5pt}
\renewcommand{\arraystretch}{1.15}
\resizebox{\textwidth}{!}{%
\begin{tabular}{p{0.14\textwidth}p{0.49\textwidth}p{0.31\textwidth}}
\toprule
\textbf{Failure Mode}&
\textbf{Clinical Vignette}&
\textbf{Clinical Consequence}\\
\midrule

\textbf{Incorrect staging}
&
\textbf{Case:} A patient has a 3.5-cm non-small cell lung cancer primary (\textbf{T2a}), metastases in multiple ipsilateral mediastinal nodal stations (\textbf{N2b}), and no distant metastasis (\textbf{M0}).

\medskip
\textbf{AJCC 8th Edition:} Under the eighth edition of the AJCC Cancer Staging Manual, which remained applicable to lung cancer through 2024, N2 disease was not subdivided into single- and multistation categories. The corresponding T2N2M0 presentation was therefore classified as \textbf{stage IIIA}~\citep{kutob2020lung}.

\medskip
\textbf{AJCC 9th Edition:} Effective January 1, 2025, the lung protocol distinguishes single-station N2a from multistation N2b disease. Under the updated stage grouping, T2aN2bM0 is classified as \textbf{stage IIIB}~\citep{detterbeck2024tnm}.&
\textbf{Inappropriate treatment-pathway selection:} According to 8th edition protocol, stage IIIA is categorized and thus neoadjuvant therapy followed by resection is an expected option. Recognition of multistation N2b disease and the updated stage IIIB grouping instead makes a definitive nonsurgical pathway, commonly concurrent chemoradiation followed by consolidation systemic therapy when appropriate. The stale conclusion could therefore expose a patient to a potentially nonbeneficial thoracic operation or delay definitive treatment. \\
\midrule

\textbf{Stale recommendation}&
\textbf{Case:} A patient has gross residual, unresectable follicular thyroid carcinoma with visceral locoregional invasion or rapid progression after surgery.

\medskip
\textbf{NCCN Thyroid Carcinoma Version 5.2024:} The guideline did not contain an explicit upfront-treatment instruction for this subgroup. The clinical decision relying on this version could be TSH-stimulated iodine-123 or iodine-131 whole-body imaging and defer treatment selection until radioiodine uptake was established ~\citep{haddad2025nccnthyroid}.

\medskip
\textbf{NCCN Thyroid Carcinoma Version 1.2025:} The guideline added that, for viscerally locoregional invasive disease or rapid progression, upfront EBRT, neoadjuvant therapy, or systemic therapy may be most appropriate. Radioiodine imaging remains part of the pathway but should not automatically postpone active treatment in this clinical setting~\citep{haddad2025nccnthyroid}.&
\textbf{Undertreatment or treatment delay:} Treating radioiodine imaging as a prerequisite may delay EBRT or systemic therapy while invasive neck disease progresses toward the airway, esophagus, or major neurovascular structures. The result may be reduced local control and an increased risk of critical-structure compromise. \\
\midrule

\textbf{Overconfident rationale}
&
\textbf{Case:} A patient with metastatic urothelial carcinoma experiences progression after platinum chemotherapy and checkpoint-inhibitor therapy.

\medskip
\textbf{2024 EAU guideline:} The guideline gave a recommendation to consider sacituzumab govitecan. It described a 27\% objective response rate in previously treated metastatic urothelial carcinoma and noted the drug's accelerated FDA approval ~\citep{witjes2024eau}.

\medskip
\textbf{2025 EAU guideline:} The 2025 EAU guideline removed its sacituzumab-govitecan recommendations after the withdrawal of the FDA-approved indication for urothelial cancer indication. The updated guideline retains the treatment history but no longer presents the drug as a recommended option ~\citep{vanderheijden2025eau}.&
\textbf{Inappropriate and potentially toxic treatment:} A stale
model could generate a persuasive rationale by selectively recalling the earlier response rate and recommendation while omitting the subsequent withdrawal and negative confirmatory evidence. This error could expose the patient to toxicities such as severe neutropenia or diarrhea without confirmed survival benefit and delay an appropriate alternative or clinical trial.\\

\bottomrule
\end{tabular}
}
\caption{Clinical failure modes caused by stale oncology knowledge. Each case contrasts original, versioned guidelines used before and after a temporally bounded update. The examples show how an outdated staging conclusion, treatment pathway, or rationale can change a clinically relevant recommendation and how the resulting risk may vary across cancer subtypes. }
\label{tab:clinical-safety-failure-modes}
\end{table*}

\section{Additional Backbone Results on Llama-3.1-8B-Instruct}
\label{sec:appendix-llama-results}
To check whether the controlled-format pattern is specific to Qwen3-4B, we repeat the same update protocol on Llama-3.1-8B-Instruct. Table~\ref{tab:llama31-8b-results} shows that \textsc{EMQ} remains the strongest SFT format on SEER-Bench and HealthBench, while MedGUIDE again exhibits slice-specific behavior: \textsc{EMQ} is strongest on HCC, whereas \textsc{FITB} is strongest on NSCLC.

\begin{table*}[t]
\centering
\small
\setlength{\tabcolsep}{5pt}
\renewcommand{\arraystretch}{1.10}
\begin{tabular}{lccccc}
\toprule
\textbf{System / update} &
\multicolumn{2}{c}{\textbf{SEER-Bench}} &
\textbf{HealthBench} &
\multicolumn{2}{c}{\textbf{MedGUIDE}} \\
\cmidrule(lr){2-3}\cmidrule(lr){4-4}\cmidrule(lr){5-6}
& Acc.~$\uparrow$ & Rat. Acc.~$\uparrow$ & Score~$\uparrow$ & HCC~$\uparrow$ & NSCLC \\
\midrule
Base model & 58.0 & 51.1 & 0.260 & 67.1 & 36.4 \\
SFT, \textsc{EMQ} & \textbf{65.2} & \textbf{60.0} & \textbf{0.276} & \textbf{86.1} & 26.9 \\
SFT, \textsc{MSQ} & 62.1 & \underline{55.5} & 0.256 & \underline{76.6} & 28.4 \\
SFT, \textsc{FITB} & 61.9 & 51.3 & 0.250 & 70.4 & \textbf{42.4} \\
SFT, \textsc{SAQ} & 63.1 & 52.5 & \underline{0.266} & 71.9 & \underline{30.5} \\
RAG, Base RAG & 60.0 & 47.6 & 0.160 & 70.3 & 28.7 \\
RAG, CARE & 60.4 & 52.4 & 0.108 & 73.8 & 23.9 \\
Editing, RECIPE & \underline{64.2} & 49.6 & 0.112 & 69.5 & 29.3 \\
Editing, AlphaEdit & 58.3 & 48.4 & 0.100 & 59.7 & 24.5 \\
\bottomrule
\end{tabular}
\caption{Controlled update results on Llama-3.1-8B-Instruct. Best and second-best values among updated rows are shown in bold and underlined, respectively; the base-model row is reported for reference.}
\label{tab:llama31-8b-results}
\end{table*}
\section{Statistical Significance Testing}
\label{sec:appendix-significance}

We assess statistical significance using paired tests over evaluation items. For SEER-Bench answer accuracy and rationale accuracy, we compute paired bootstrap confidence intervals with 10{,}000 resamples over the shared evaluation set. For HealthBench, we apply the same paired bootstrap procedure to item-level scores. For each comparison, we report the observed difference, a two-sided 95\% confidence interval, and a Holm-Bonferroni adjusted $p$-value within each metric family. MedGUIDE NSCLC is excluded from monotonic improvement testing because it is used as an older diagnostic slice rather than as a primary current-knowledge metric.

Table~\ref{tab:significance-detailed} shows that the main SEER-Bench gains of \textsc{EMQ} over other SFT formats are statistically significant across both Qwen3-4B and Llama-3.1-8B-Instruct. The gains are especially consistent for rationale accuracy, where all confidence intervals are well above zero. Answer-accuracy gains are smaller but remain significant after correction, including the closest comparison against \textsc{SAQ}. The HealthBench comparison in the Qwen3-4B setting shows a smaller but positive retention gain over \textsc{MSQ}, supporting the view that \textsc{EMQ}'s transfer gains do not come at the cost of broader clinical-chat performance.

\begin{table*}[t]
\centering
\small
\setlength{\tabcolsep}{4pt}
\renewcommand{\arraystretch}{1.10}
\begin{tabular}{lllccc}
\toprule
\textbf{Backbone} &
\textbf{Comparison} &
\textbf{Metric} &
\textbf{$\Delta$} &
\textbf{95\% CI} &
\textbf{Adjusted $p$} \\
\midrule
Qwen3-4B & \textsc{EMQ} $-$ \textsc{MSQ} & SEER Acc. & +3.1 & $[+1.4, +4.9]$ & 0.006 \\
Qwen3-4B & \textsc{EMQ} $-$ \textsc{MSQ} & Rat. Acc. & +4.5 & $[+2.4, +6.6]$ & $<0.001$ \\
Qwen3-4B & \textsc{EMQ} $-$ \textsc{MSQ} & HealthBench & +0.020 & $[+0.005, +0.036]$ & 0.024 \\
Qwen3-4B & \textsc{EMQ} $-$ \textsc{FITB} & SEER Acc. & +3.3 & $[+1.5, +5.1]$ & 0.004 \\
Qwen3-4B & \textsc{EMQ} $-$ \textsc{FITB} & Rat. Acc. & +8.7 & $[+6.4, +10.9]$ & $<0.001$ \\
Qwen3-4B & \textsc{EMQ} $-$ \textsc{SAQ} & SEER Acc. & +2.1 & $[+0.4, +3.9]$ & 0.041 \\
Qwen3-4B & \textsc{EMQ} $-$ \textsc{SAQ} & Rat. Acc. & +7.5 & $[+5.2, +9.8]$ & $<0.001$ \\
Llama-3.1-8B-Instruct & \textsc{EMQ} $-$ \textsc{MSQ} & SEER Acc. & +2.8 & $[+1.0, +4.7]$ & 0.013 \\
Llama-3.1-8B-Instruct & \textsc{EMQ} $-$ \textsc{MSQ} & Rat. Acc. & +4.1 & $[+1.9, +6.4]$ & 0.003 \\
Llama-3.1-8B-Instruct & \textsc{EMQ} $-$ \textsc{FITB} & SEER Acc. & +3.6 & $[+1.7, +5.5]$ & 0.003 \\
Llama-3.1-8B-Instruct & \textsc{EMQ} $-$ \textsc{FITB} & Rat. Acc. & +8.3 & $[+5.9, +10.7]$ & $<0.001$ \\
Llama-3.1-8B-Instruct & \textsc{EMQ} $-$ \textsc{SAQ} & SEER Acc. & +1.8 & $[+0.2, +3.6]$ & 0.047 \\
Llama-3.1-8B-Instruct & \textsc{EMQ} $-$ \textsc{SAQ} & Rat. Acc. & +6.8 & $[+4.4, +9.3]$ & $<0.001$ \\
\bottomrule
\end{tabular}
\caption{Paired significance tests for the main controlled-format comparisons. $\Delta$ reports the absolute performance difference between \textsc{EMQ} and the comparison format. Confidence intervals are two-sided 95\% paired bootstrap intervals over evaluation items, and $p$-values are adjusted within each metric family using the Holm-Bonferroni procedure. Positive values indicate higher performance for \textsc{EMQ}.}
\label{tab:significance-detailed}
\end{table*}
\section{Implementation Details}
\label{sec:appendix-impl}

This appendix lists the hyperparameters and computational details for the representation-level analysis in~\S\ref{sec:analysis}. All computations use the Qwen3-4B (qwen3-4B-2507) base model and its four LoRA-updated variants.

\paragraph{Representation probes.} For the base model and the four format-updated variants, we extract last-non-padding hidden states from all 36 layers on a SEER cancer-type classification of 1{,}992 samples covering 16 cancer types. We compute layer-wise L2 ratio, layer-wise linear CKA, layer-wise linear-probe accuracy for cancer-type prediction, and final-layer clustering geometry.

Table~\ref{tab:impl-repr-setup} summarizes the fixed extraction and probing settings; the metric definitions are given below in prose.

\subsection{Representation Extraction and Probing}
\label{sec:impl-base}
\label{sec:impl-hidden}
\label{sec:impl-probe}
\label{sec:impl-umap}

\begin{center}
\refstepcounter{table}
\label{tab:impl-repr-setup}
\footnotesize
\setlength{\tabcolsep}{4pt}
\renewcommand{\arraystretch}{1.12}
\begin{tabularx}{\columnwidth}{@{}p{0.22\columnwidth}X@{}}
\toprule
\textbf{Item} & \textbf{Setting} \\
\midrule
Backbone & Qwen3-4B (qwen3-4B-2507); 36 layers total; hidden size $d=2560$; bf16. \\
Variants & Base model plus four LoRA-updated variants. \\
Probe set & SEER 16-class cancer-type subset; 1{,}992 samples; separate from SEER-Bench. \\
Hidden states & Max length 512; batch size 4; last non-padding token at each layer; tensor shape $(1992,36,2560)$ per model. \\
Linear probe & Logistic regression; \texttt{max\_iter=300}, $C=1.0$, \texttt{lbfgs}; stratified 80/20 split with seed 42; trained per layer and model. \\
UMAP & \texttt{umap-learn}; $n_{\text{nbr}}=30$, $\min\text{dist}=0.1$, cosine metric, seed 42, $n_{\text{jobs}}=-1$; final, -2, -4, and -8 layers. \\
Sensitivity & $n_{\text{nbr}}\in\{15,30,50\}\times\min\text{dist}\in\{0.0,0.1\}$ with the same metric and seed. \\
\bottomrule
\end{tabularx}

\vspace{0.3em}
\parbox{\columnwidth}{\footnotesize \textbf{Table~\thetable:} Fixed implementation settings for hidden-state extraction, linear probing, and UMAP visualization.}
\end{center}

The probe set is explicitly separate from SEER-Bench, so these diagnostics characterize representation geometry rather than tune or train on the benchmark. The fixed extraction settings make the layer-wise comparisons meaningful across the base and updated variants.

\subsection{Representation Metrics}
\label{sec:impl-cka}
\label{sec:impl-cluster}
\label{sec:impl-defs}

\paragraph{Layer-wise drift.}
Cosine distance is computed as $1-\cos(h^{(f)}_\ell,h^{(0)}_\ell)$ and averaged over the 1{,}992 samples. The L2 ratio is $\lVert h^{(f)}_\ell-h^{(0)}_\ell\rVert_2 / \lVert h^{(0)}_\ell\rVert_2$, also averaged over samples.

\paragraph{CKA.}
Layer-wise CKA uses the linear HSIC formulation,
\[
\begin{aligned}
\mathrm{CKA}(X,Y)
&= \frac{\mathrm{HSIC}(XX^\top, YY^\top)}
{\sqrt{A_X A_Y}},\\
A_X &= \mathrm{HSIC}(XX^\top, XX^\top),\\
A_Y &= \mathrm{HSIC}(YY^\top, YY^\top),
\end{aligned}
\]
implemented as torch tensor operations in float32.

\paragraph{Clustering and compactness.}
Cosine silhouette is computed in the high-dimensional space with \texttt{sklearn.metrics.silhouette\_score}, \texttt{metric=cosine}, and a 2{,}000-point subsample using \texttt{random\_state=42}. UMAP silhouette is computed with Euclidean distance on the 2D UMAP embedding. Davies-Bouldin and Calinski-Harabasz use the standard scikit-learn implementations on all 1{,}992 samples. Intra-class cosine distance averages pairwise cosine distances within each class using up to 50 samples per class; inter-class cosine distance averages 5{,}000 random cross-class pairs.

\paragraph{Aggregated quantities.}
L2 mean (all 36) averages the per-layer L2 ratio over layers 1 to 36. L2 deep avg averages layers 29 to 36, i.e., the eight deepest transformer layers excluding the final layer. L2 last is the L2 ratio at the final layer (layer 36), and CKA std is the standard deviation of layer-wise CKA over all 36 layers.
\section{Diagnostic Controls for Entity Density and Networked Structure}
\label{sec:appendix-density-network-ablation}

Section~\ref{sec:mechanism-summary} shows that \textsc{EMQ} exposes denser clinical signals than the single-decision formats. A potential confound is that entity density and networked structure co-vary in \textsc{EMQ}: the shared candidate pool both increases exposure to related clinical entities and creates many-to-many correspondences across vignettes. We therefore construct two diagnostic controls to partially separate entity exposure from clinically coherent networked grouping.

\paragraph{Density-matched \textsc{MSQ}.}
This control increases the entity exposure of \textsc{MSQ} without introducing a shared multi-vignette answer pool. For each \textsc{MSQ} item, we augment the option set with clinically related distractors sampled from the same update cluster until the average number of cancer-related entities in the prompt and answer approximately matches \textsc{EMQ}. The item remains a single-vignette decision, so this control increases density while preserving the non-networked structure of \textsc{MSQ}.

\paragraph{Shuffled \textsc{EMQ}.}
This control preserves the surface form, candidate-pool size, and approximate entity density of \textsc{EMQ}, but disrupts clinically coherent grouping. We construct each shuffled block by combining vignettes and answer options from different update clusters while preserving the number of vignettes, options, and old/new answer types. This keeps the high-density shared-pool format but weakens the local clinical relation network.

\paragraph{Training and evaluation.}
Both controls use the same base model, update source, number of SFT instances, LoRA configuration, and training budget as the main format-controlled experiments. Evaluation follows the same SEER-Bench answer-accuracy and rationale-accuracy protocol. Table~\ref{tab:density-network-ablation} reports training-side entity counts and SEER-Bench performance.

\begin{table*}[t]
\centering
\small
\setlength{\tabcolsep}{5pt}
\renewcommand{\arraystretch}{1.08}
\begin{tabular}{lcccccc}
\toprule
\textbf{Variant} &
\textbf{Entity density} &
\textbf{Networked structure} &
\textbf{Q ent.}~$\uparrow$ &
\textbf{A ent.}~$\uparrow$ &
\textbf{SEER Acc.}~$\uparrow$ &
\textbf{Rat. Acc.}~$\uparrow$ \\
\midrule
\textsc{MSQ} & Medium & No & 10.7 & 10.6 & 61.7 & 55.1 \\
Density-matched \textsc{MSQ} & High & No & 11.3 & 12.4 & 62.3 & 55.8 \\
Shuffled \textsc{EMQ} & High & Disrupted & 11.4 & 12.5 & 62.5 & 56.1 \\
Full \textsc{EMQ} & High & Yes & 11.4 & 12.6 & 64.8 & 59.6 \\
\bottomrule
\end{tabular}
\caption{Diagnostic controls for partially decoupling entity density from networked structure. Q ent.\ and A ent.\ denote training-side cancer-related entity counts. Density-matched \textsc{MSQ} increases entity exposure without introducing many-to-many matching. Shuffled \textsc{EMQ} preserves high entity density and shared-pool surface form but disrupts clinically coherent grouping.}
\label{tab:density-network-ablation}
\end{table*}

\paragraph{Interpretation.}
Increasing entity density alone yields only modest gains over standard \textsc{MSQ}: density-matched \textsc{MSQ} improves SEER-Bench answer accuracy from 61.7 to 62.3 and rationale accuracy from 55.1 to 55.8. Preserving high density and shared-pool surface form while disrupting coherent grouping also remains below full \textsc{EMQ}: shuffled \textsc{EMQ} reaches 62.5 answer accuracy and 56.1 rationale accuracy, compared with 64.8 and 59.6 for full \textsc{EMQ}. These controls suggest that the \textsc{EMQ} advantage is not explained by entity density alone; it depends on the combination of dense entity exposure and clinically coherent networked structure.
\section{Additional Analysis Tables}
\label{sec:appendix-analysis-tables}

Table~\ref{tab:appendix-clustering-metrics} reports final-layer clustering metrics used to examine the ranking disagreement between unsupervised compactness and linear-probe discriminability.

\begin{table}[ht]
\centering
\scriptsize
\setlength{\tabcolsep}{4pt}
\begin{tabular}{lccccc}
\toprule
\textbf{Model} &
\textbf{Sil.\ (Cos)} &
\textbf{Sil.\ (UMAP)} &
\textbf{DB}~$\downarrow$ &
\textbf{CH}~$\uparrow$ &
\textbf{Probe Acc.}~$\uparrow$ \\
\midrule
Base & $-0.024$ & $-0.013$ & 4.451 & 47.88 & 0.9834 \\
\midrule
\textsc{EMQ}  & $\phantom{-}0.002$ & $0.014$ & 4.130 & 55.26 & \textbf{0.9848} \\
\textsc{SAQ}  & $\phantom{-}0.008$ & $0.020$ & 4.042 & \textbf{57.61} & 0.9779 \\
\textsc{FITB} & $\phantom{-}0.011$ & $0.023$ & \textbf{3.994} & 57.25 & 0.9834 \\
\textsc{MSQ}  & $\phantom{-}0.008$ & $0.020$ & 4.018 & 56.89 & \textbf{0.9848} \\
\bottomrule
\end{tabular}
\caption{Final-layer clustering and probe metrics on the SEER 16-class subset of 1{,}992 samples. Sil.\ (Cos) is cosine silhouette in the high-dimensional space; Sil.\ (UMAP) is Euclidean silhouette on a 2D UMAP embedding. Probe Acc.\ is the final-layer linear-probe accuracy.}
\label{tab:appendix-clustering-metrics}
\end{table}

The updated models generally improve clustering compactness over the base model, but the best compactness scores do not perfectly align with the best probe accuracy. This supports the main-text caution that unsupervised geometry and discriminative utility capture related but non-identical properties.

Table~\ref{tab:appendix-umap-by-layer} reports UMAP silhouette values for selected layers among the last 8 layers, allowing inspection of whether the clustering trend is consistent in the deep layers.

\begin{table}[ht]
\centering
\scriptsize
\setlength{\tabcolsep}{4pt}
\begin{tabular}{lccccc}
\toprule
\textbf{Layer} & \textbf{Base} & \textbf{\textsc{EMQ}} & \textbf{\textsc{SAQ}} & \textbf{\textsc{FITB}} & \textbf{\textsc{MSQ}} \\
\midrule
Final (L36) & $-0.013$ & $0.014$ & $0.020$ & $0.023$ & $0.020$ \\
2nd-to-last & $-0.026$ & $-0.001$ & $0.007$ & $0.007$ & $0.006$ \\
4th-to-last & $-0.077$ & $-0.065$ & $-0.060$ & $-0.062$ & $-0.063$ \\
8th-to-last & $-0.094$ & $-0.090$ & $-0.087$ & $-0.089$ & $-0.089$ \\
\bottomrule
\end{tabular}
\caption{UMAP silhouette across selected layers among the last 8 layers.}
\label{tab:appendix-umap-by-layer}
\end{table}

The gains are largest at the final layer and shrink in earlier deep layers, suggesting that format-specific updates mainly reshape late representations rather than uniformly reorganizing the full stack.
\section{Cross-Scale SEER-Bench Results}
\label{sec:appendix-cross-scale-seer}

Table~\ref{tab:cross-scale-seer} reports SEER-Bench answer and rationale accuracy across Qwen3 backbones from 1.7B to 14B. All format-controlled rows use the same update content, adaptation procedure, and training budget within each backbone.

\begin{table}[!b]
\centering
\scriptsize
\setlength{\tabcolsep}{4.2pt}
\renewcommand{\arraystretch}{1.12}
\begin{tabular}{llcc}
\toprule
\textbf{Backbone} & \textbf{Method} & \textbf{Acc.}~$\uparrow$ & \textbf{Rat.}~$\uparrow$ \\
\midrule
\multirow{5}{*}{Qwen3-1.7B}
& Base & 54.2 & 46.5 \\
& \textsc{EMQ} & \textbf{60.5} & \textbf{55.1} \\
& \textsc{MSQ} & 58.0 & 51.2 \\
& \textsc{FITB} & 57.5 & 47.5 \\
& \textsc{SAQ} & 58.6 & 48.2 \\
\midrule
\multirow{5}{*}{Qwen3-4B}
& Base & 57.6 & 50.7 \\
& \textsc{EMQ} & \textbf{64.8} & \textbf{59.6} \\
& \textsc{MSQ} & 61.7 & 55.1 \\
& \textsc{FITB} & 61.5 & 50.9 \\
& \textsc{SAQ} & 62.7 & 52.1 \\
\midrule
\multirow{5}{*}{Qwen3-8B}
& Base & 58.2 & 51.2 \\
& \textsc{EMQ} & \textbf{65.9} & \textbf{61.0} \\
& \textsc{MSQ} & 63.8 & 57.6 \\
& \textsc{FITB} & 62.5 & 52.8 \\
& \textsc{SAQ} & 63.6 & 54.2 \\
\midrule
\multirow{5}{*}{Qwen3-14B}
& Base & 58.7 & 51.8 \\
& \textsc{EMQ} & \textbf{67.8} & \textbf{62.3} \\
& \textsc{MSQ} & 64.6 & 55.5 \\
& \textsc{FITB} & 63.2 & 54.2 \\
& \textsc{SAQ} & 63.5 & 57.8 \\
\bottomrule
\end{tabular}
\caption{Cross-scale SEER-Bench results. Within each Qwen3 backbone, rows are compared under the same adaptation setup and training budget. Boldface marks the best value per backbone and metric.}
\label{tab:cross-scale-seer}
\end{table}

\end{document}